\documentclass[journal]{IEEEtran}
\usepackage{amsmath,amsfonts}
\usepackage{algorithmic}
\usepackage{algorithm}
\usepackage{array}
\usepackage{amssymb}
\usepackage{amsmath}
\usepackage{tabularx}
\usepackage{amsthm}
\usepackage{amsmath}
\usepackage{pifont}
\usepackage{textcomp}
\usepackage{stfloats}
\usepackage{multirow,booktabs,color,soul,threeparttable}
\usepackage[table]{xcolor}
\usepackage{booktabs}
\usepackage[colorlinks]{hyperref}

\usepackage{bigstrut,multirow,rotating}
\usepackage{longtable}
\usepackage{subfigure} 
\usepackage{url}
\usepackage{verbatim}
\usepackage{graphicx}
\usepackage{cite}
\begin{document}

\title{Variable Search Stepsize for Randomized Local Search in Multi-Objective Combinatorial Optimization}

\author{IEEE Publication Technology,~\IEEEmembership{Staff,~IEEE,}
\author{Xuepeng Ren, Maocai Wang, Guangming Dai, Zimin Liang, Qianrong Liu,\\ Shengxiang Yang~\IEEEmembership{Fellow,~IEEE}, Miqing Li~\IEEEmembership{Senior Member,~IEEE} 
\thanks{This work was supported by National Natural Science Foundation of China under Grant No.~42571407 \& No.~42271391, Hubei Excellent Young and Middle-aged Science and Technology Innovation Team Plan Project under Grant No.~T2021031, and Royal Society International Exchanges 2025 Cost Share (NSFC) under Grant No.~IEC\textbackslash{NSFC}\textbackslash252114 (Corresponding author: Maocai Wang, Shengxiang Yang and Miqing Li).}
\thanks{Xuepeng Ren, Maocai Wang and Guangming Dai are with School of Computer, China University of Geosciences, Engineering Research Center of Natural Resource Information Management and Digital Twin Engineering Software, Ministry of Education, and Hubei Key Laboratory of Intelligent Geo-Information Processing, Wuhan 430074, China. (e-mail: renxuepeng@cug.edu.cn, cugwangmc@126.com and gmdai@cug.edu.cn).}
\thanks{Shengxiang Yang and Xuepeng Ren are with the School of Computer, Science and Informatics, De Montfort University, Leicester LE1 9BH, U.K. (e-mail: syang@dmu.ac.uk).}
\thanks{Zimin Liang and Miqing Li are with the School of Computer Science, University of Birmingham, Birmingham B15 2TT, U.K. (e-mail: zxl525@student.bham.ac.uk and m.li.8@bham.ac.uk).}
\thanks{Qianrong Liu is with the School of Chemical Engineering, University of Birmingham, Birmingham B15 2TT, U.K. (e-mail: qxl242@student.bham.ac.uk).}
}
}

\markboth{}%
{Ren \MakeLowercase{\textit{et al.}}: Variable Search Stepsize for Randomized Local Search in Multi-Objective Combinatorial Optimization}
\maketitle

\begin{abstract}
Over the past two decades, research in evolutionary multi-objective optimization has predominantly focused on continuous domains, with comparatively limited attention given to multi-objective combinatorial optimization problems (MOCOPs). Combinatorial problems differ significantly from continuous ones in terms of problem structure and landscape. Recent studies have shown that on MOCOPs multi-objective evolutionary algorithms (MOEAs) can even be outperformed by simple randomised local search. Starting with a randomly sampled solution in search space, randomised local search iteratively draws a random solution (from an archive) to perform local variation within its neighbourhood. However, in most existing methods, the local variation relies on a fixed neighbourhood, which limits exploration and makes the search easy to get trapped in local optima. In this paper, we present a simple yet effective local search method, called variable stepsize randomized local search (VS-RLS), which adjusts the stepsize during the search. VS-RLS transitions gradually from a broad, exploratory search in the early phases to a more focused, fine-grained search as the search progresses. We demonstrate the effectiveness and generalizability of VS-RLS through extensive evaluations against local search and MOEAs methods on diverse MOCOPs.
\end{abstract}


\section{INTRODUCTION}
\IEEEPARstart{M}ulti-objective combinatorial optimization problems (MOCOPs) aim to optimize two or more conflicting objectives at the same time, and their solution space is made up of discrete decision variables \cite{count1}. These problems are common in many fields, such as software engineering \cite{chen2023weights,hierons2020many}, logistics scheduling \cite{count2}, cloud computing \cite{zhu2015evolutionary} and aerospace design \cite{count3}. Multi-objective evolutionary algorithms (MOEAs) and local search heuristics are among popular techniques to tackle MOCOPs.

With its success in multi-objective optimization, MOEAs have become a mainstream method for solving MOCOPs. Representative algorithms include NSGA-II \cite{count7}, MOEA/D \cite{count8}, and SMS-EMOA \cite{count9}, which have been widely applied to various problems, such as the multi-objective knapsack and traveling salesman problems \cite{count12,count13,count14}. However, empirical studies have shown that MOEAs exhibit more local search behaviour than local search methods when applied to MOCOPs \cite{count6,li2025combinatorial}. Meanwhile, other empirical evidence suggests that, on certain MOCOPs, MOEAs can even be outperformed by local search algorithms \cite{count34,zimin2}.

Local search algorithms usually maintain an archive of non-dominated solutions. In each iteration, one solution is selected from the archive and local variation is applied within its neighbourhood. Early studies of local search mainly focused on scalarization-based methods, which transform a multi-objective problem into multiple single-objective problems in different search directions \cite{SCAandPar}. With the introduction of Pareto local search, researchers began to use Pareto dominance directly to guide the search, where two main types of methods have been proposed \cite{count26,liang2026random}: systematic and randomized local search. In systematic local search, candidate solutions in the neighbourhood are examined according to a predetermined order, for example by fully scanning the neighbourhood or using a first-improvement strategy \cite{PLS,PLS23}. In randomized local search, one parent solution is randomly selected from the archive, and an offspring solution is generated within its neighbourhood (e.g., by bit-flip), after which the archive is updated according to the Pareto dominance relation \cite{2Relatedwork14}. 

Compared with the systematic approach, randomized local search often produces higher-quality solution sets for MOCOPs \cite{count34,liang2026random}. Most existing randomized local search algorithms rely on a fixed neighbourhood structure to generate candidate solutions \cite{2Relatedwork14,1Relatedwork1,PLS17,1GSEMO,BCGSEMO}. However, the fixed setting may restrict the exploration ability, leading to early stagnation of the search. For example, when flipping one decision variable at a time, the local search may get trapped in local optima.
On the other hand, flipping many variables simultaneously can lead to large jumps that can disrupt fine-grained improvements. 

To address this issue, we propose a simple randomized local search method called Variable Stepsize Randomized Local Search (VS-RLS). In VS-RLS, the stepsize refers to the neighbourhood used to generate offspring. For example, in binary problems, a stepsize of one corresponds to the 1-bit neighbourhood, whereas in permutation problems it corresponds to a neighbourhood defined by the same number of exchange or rearrangement operations (e.g., 2-opt in TSP).  In each iteration, the search starts from the neighbourhood with a stepsize of one and progressively expands to larger stepsize (i.e., two, three, and so on) until a good solution (i.e., a solution that is not dominated by any solution in the archive) is found or the maximum stepsize is reached. For each stepsize, we do not perform a full neighbourhood scan; instead we conduct some sampling. This avoids an exhaustive search whose computational cost will increase exponentially with the stepsize, while allowing us to quickly identify promising regions on which the search can focus in later stages.
The proposed method is verified on a wide range of multi-objective combinatorial problems, including the multi-objective knapsack \cite{KP}, travelling salesman \cite{TSP}, quadratic assignment \cite{QAP}, and NK-landscape \cite{NK} problems.

The remainder of this paper is organized as follows. Section II introduces the preliminaries and related work. Section III details the proposed algorithm. Section IV outlines the experimental design. Section V presents the experimental results and related discussions. Section VI concludes with key findings and future research directions. All codes and experimental data have been made available for reproducibility: \url{https://github.com/vsrlsanonymous/VS-RLS-code}.

\section{PRELIMINARIES AND RELATED WORK}
In this section, we first introduce the definitions of MOCOPs and local search, and then review in detail the related work on MOEAs and local search in the context of MOCOPs.

\subsection{Definitions}
MOCOPs involve optimizing multiple conflicting objectives simultaneously, where decision variables are discrete and the search space is large and complex. Consider a minimization MOCOP with $m$ objective functions:
\begin{equation}
f(x) = (f_1(x), f_2(x), \dots, f_m(x)),
\end{equation}
where $x$ is a solution in the finite decision space $X$. 
Depending on the problem, $X$ can consist of binary, integer, or permutation representations. 
$D$ denotes the dimensionality of the decision vector, and $f(x)$ maps $x$ to the objective space $Z$, with $Z \subseteq \mathbb{R}^m$.

The dominance relationship between two objective vectors $z, z' \in Z$ is defined as:
\begin{equation}
z \prec z' \iff \left( z_i \leq z'_i , \forall i \in \{1, \dots, m\} \right) \wedge \left( z \neq z' \right),
\end{equation}
meaning that $z$ is no worse in all objectives and strictly better in at least one.
For two solutions $x, x' \in X$, dominance is extended as
\begin{equation}
x \prec x' \iff f(x) \prec f(x').
\end{equation}
A solution $x \in X$ is said to be non-dominated with respect to a set $\mathcal{A} \subseteq X$ 
if no $x' \in \mathcal{A}$ satisfies $x' \prec x$, i.e., $\nexists x' \in \mathcal{A}: x' \prec x$.

Local search is a fundamental optimization strategy that iteratively improves a candidate solution by exploring its neighbourhood. 
Given a solution \(x \in X\), its neighbourhood \(N(x)\) can be defined in different ways. 
For binary problems, a common choice is the Hamming neighbourhood:
\begin{equation}
N_{\mathrm{H}}(x) = \{\,x' \in X \mid d_{\mathrm{H}}(x,x') = r \,\},\quad r\geq1,
\end{equation}
where \(d_{\mathrm{H}}(x,x')\) denotes the Hamming distance between \(x\) and \(x'\). 
The case \(r=1\) corresponds to the standard one-bit-flip neighbourhood, while larger \(r\) values correspond to multiple-bit-flips. 
More generally, neighbourhoods depend on the solution representation. 
For example, in permutation-based problems such as the traveling salesman problem, 
the neighbourhood is often constructed by the 2-opt operator \cite{canshu}, 
while in the quadratic assignment problem, the 2-swap operator \cite{canshu} is widely used. 

In single-objective optimization, a solution is called a local optimum if no neighbour improves its objective value. 
In the multi-objective case, this concept is extended to Pareto local optimality. 
A Pareto local optimum (PLO) is a solution \(x \in X\) such that no neighbour dominates it:
\begin{equation}
x \text{ is a PLO} \iff \nexists x' \in N(x) \; \text{such that} \; f(x') \prec f(x).
\end{equation}

\subsection{MOEAs in MOCOPs}
MOEAs achieve effective approximations of the Pareto front by exploring diverse solution sets, and they have demonstrated strong performance in continuous optimization problems \cite{count10}. However, when the focus shifts to MOCOPs, their performance is often limited \cite{count6}. The discrete decision variables and highly scattered search landscapes of MOCOPs introduce new challenges for classical MOEAs in terms of search efficiency and solution quality \cite{li2025combinatorial}.

Despite these challenges, MOEAs remain one of the major approaches for solving MOCOPs. Existing studies enhance their performance by designing problem-specific encoding schemes, variation operators and population update mechanisms \cite{count12,count13,count14,liang2023non,chu2024improving}, which have led to successful applications on typical combinatorial problems such as multi-objective knapsack and travelling salesman. MOEAs have also shown good adaptability in many real-world applications \cite{count15,count16,count17,NAS1,NAS2,NAS3,count20}.

To further improve the search efficiency on MOCOPs, hybrid frameworks that integrate MOEAs with local improvement techniques have attracted increasing attention. Some studies incorporate local search as an internal component of the evolutionary process \cite{count23}, while others apply it as a post-processing step to enhance the obtained solution set \cite{1Relatedwork4}.

Note that in the area of runtime analysis, there have been many theoretical studies analyzing MOEAs on MOCOPs, such as on subset selection~\cite{Qian2025, deng_runtime_2024, bian_robust_2022} and minimum spanning tree~\cite{do_rigorous_2023, neumann_expected_2007}.
In particular, artificial MOCOPs (i.e., pseudo-Boolean problems~\cite{Liang2025}) are the most commonly studied problems for the runtime analysis of mainstream MOEAs~\cite{zheng_first_2022, doerr_first_2023, zheng_how_2024, doerr_proven_2024}, the development of MOEAs (e.g., considering stochasticity~\cite{bian_stochastic_2025, ren2025stochastic}), algorithmic components (e.g., archiving \cite{bian2024archive,ren2026not} and different variation operators of MOEAs~\cite{Bit1, NSEMO6}, where search algorithms may perform differently from practical MOCOPs \cite{liang2026random}.

\subsection{Local Search in MOCOPs}
Local search has been widely applied to MOCOPs because of its simplicity and efficiency in exploring discrete solution spaces \cite{count25,count27,count28,count29,liang2026random}. Existing multi-objective local search methods can be divided into scalarization-based and Pareto-based approaches \cite{SCAandPar}. Early scalarization methods decompose a multi-objective problem into multiple single-objective subproblems and then apply classical search strategies \cite{RW44}. In contrast, Pareto-based methods directly use dominance relations to guide the search and have become the mainstream in multi-objective combinatorial optimization \cite{PLS,DMOLS2,DMOLS4}.

Pareto-based local search methods can be further divided into systematic and randomized strategies \cite{count26}. Systematic methods aim to exhaustively explore the neighbourhood of each solution or to find a sufficiently good neighbor. Recent studies have improved neighbourhood exploration to enhance search quality and avoid premature convergence \cite{PLS22,PLS23,PLS8,PLS9}. In contrast, randomized methods sample the neighbourhood randomly based on certain criteria \cite{RW2}.

Arguably, one of the most representative randomized local search algorithms is SEMO \cite{count35}, which randomly selects a solution from the archive, applies a bit-flip variation (with a fixed neighbourhood size), and then updates the archive based on the dominance relation. Although recent studies show the effectiveness on MOCOPs \cite{count6,liang2026random}, its fixed neighbourhood mechanism limits the algorithm to small moves. As a result, it may fail to escape regions that require simultaneous multi-bit improvements \cite{Bit1}. To help the algorithm escape from local optima, some studies introduced appropriate perturbation when stagnation occurs, e.g., \cite{NSEMO10}, while others considered restart operators, e.g., \cite{NSEMO11}.

\section{PROPOSED ALGORITHM}
Although randomized local search methods have demonstrated good performance in MOCOPs, those employing a fixed stepsize may be prone to local optima. In this section, we first explain the limitations of fixed neighbourhood local search, then introduce the variable stepsize mechanism of VS-RLS, and finally outline the overall framework the proposed method.

\begin{figure}[t]
	\centering  
		\includegraphics[width=1.8in]{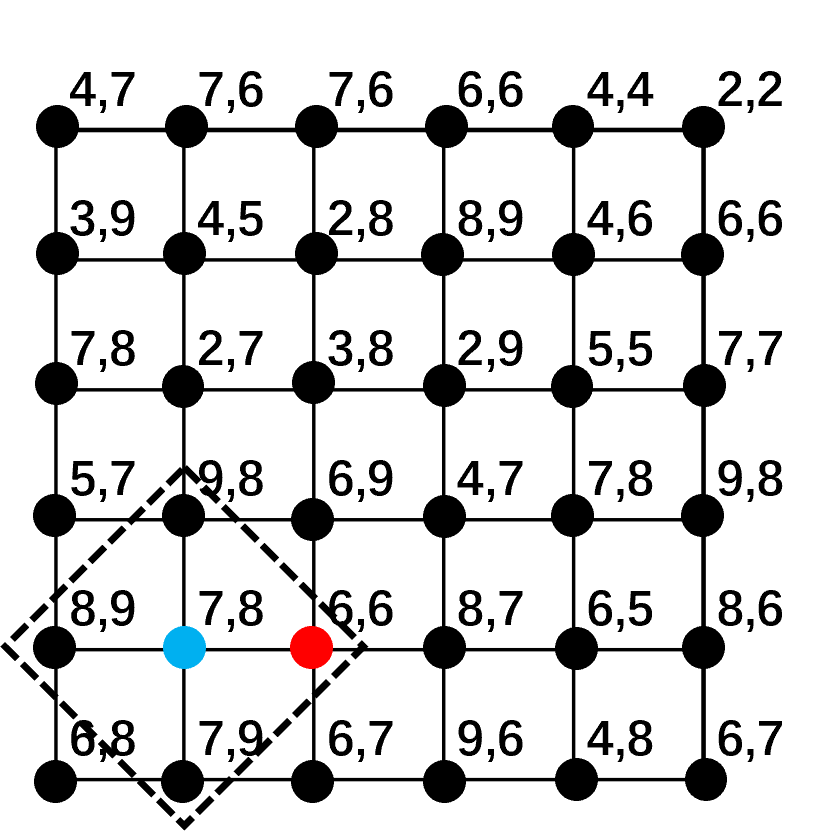} 
	\caption{An illustration of the failure of fixed stepsize randomized local search on a bi-objective minimization toy example (adapted from \cite{III1}). Each vertex represents a solution, and the numbers in the upper-right corner indicate its objective values. An edge connects two solutions whose decision variables differ by one bit. The blue point represents the starting location of the search, while the dashed rectangle indicates the search region when the stepsize is fixed to one (i.e., 1-bit neighbourhood). When performing randomized local search with a stepsize of one, one of the four neighbouring solutions around the blue point is randomly selected as an offspring. The red point represents a solution that is accepted into the archive according to the dominance relation. As can be seen, once the search reaches the local optimum at $(6,6)$ (red point), all neighbouring solutions within the fixed stepsize are dominated by it, and the algorithm is therefore unable to move forward.}
	\label{F1}
\end{figure}

\subsection{Randomized Local Search with Fixed Stepsize}

To illustrate the limitation of a fixed stepsize randomized local search, we consider a bi-objective toy example, as shown in Fig.~\ref{F1}. Each vertex represents a solution in the search space, and two neighbouring vertices differ by one bit in their decision variables. When the stepsize is fixed to one, the search is restricted to the corresponding 1-bit neighbourhood of the current solution, as indicated by the dashed rectangle. In this setting, a randomized local search generates an offspring by randomly selecting one neighbouring solution. If the offspring dominates the current solution or is non-dominated with respect to it, it is accepted into the archive according to the dominance relation. As shown in the figure, once the search reaches the local optimum at $(6,6)$, all neighbouring solutions within the fixed stepsize are dominated by the current solution. Consequently, no further progress can be made, and the search becomes trapped in this local region.


\begin{figure*}[t]
\centering  
\includegraphics[width=6.5in]{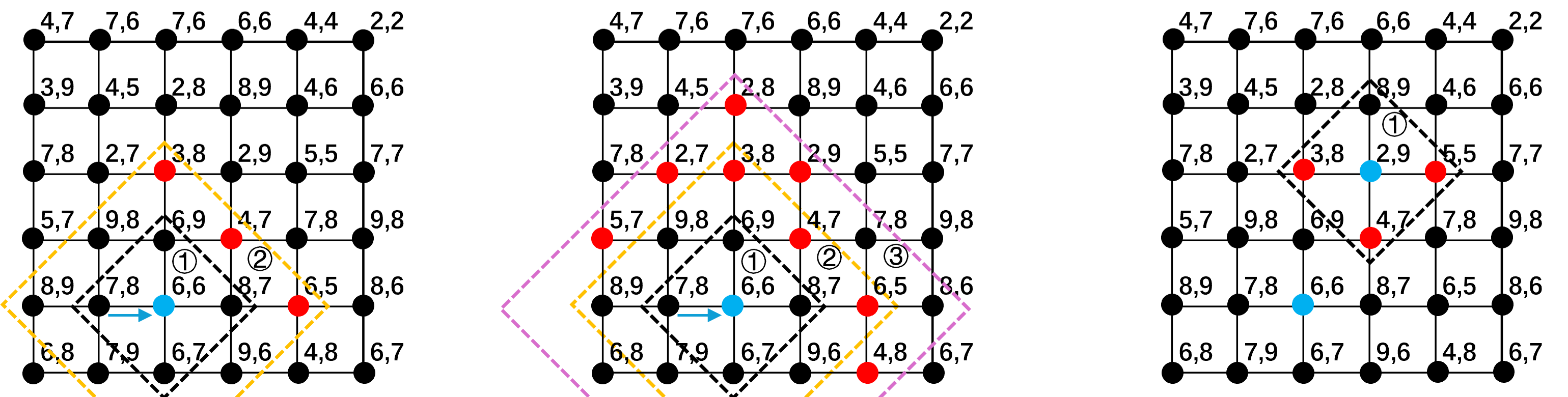} 
\caption{An illustration of the proposed VS-RLS on the bi-objective toy example in  Fig.~\ref{F1}. The dashed rectangles in different colours indicate the search regions corresponding to different stepsize, where the yellow and pink rectangles represent the search regions with stepsize two and three, respectively. Specifically, in the first subfigure, assume that the algorithm is trapped at a local optimum $(6,6)$ when the stepsize is equal to one. VS-RLS first randomly samples one solution from the search region (black dashed rectangle) corresponding to stepsize of one. Since the sampled solution is not acceptable for the archive (i.e., being dominated by at least one solution in the archive), the stepsize is increased to two (yellow dashed rectangle) and another solution is randomly sampled. Although multiple acceptable solutions exist in the search region with stepsize two, such as $(3,8)$, $(4,7)$, and $(6,5)$, suppose that none of these acceptable solutions is sampled. The stepsize is therefore further increased to three, as shown in the second subfigure. This process illustrates that, even if no acceptable solution is sampled under small stepsize, expanding the stepsize allows the algorithm to sample solutions from a broader region and thus escape from the local optimum. At the same time, since the stepsize is increased progressively from small to large values, VS-RLS preserves the ability to sample acceptable solutions near the current solution. Once an acceptable solution is sampled, the current iteration is terminated. For example, in the third subfigure, suppose that the solution $(2,9)$ is sampled when the stepsize is equal to three. Since $(2,9)$ is non-dominated with respect to $(6,6)$, it is added to the archive. The algorithm then starts a new iteration by randomly selecting a solution from the archive as the parent. In this case, if $(2,9)$ is selected as the parent, its stepsize of one region may be more likely to contain acceptable solutions than that of $(6,6)$, allowing the search to move from a less promising region to a more promising one.}
\label{F3} 
\end{figure*}

\subsection{The Proposed Variable Size Randomized Local Search (VS-RLS)}
To overcome the limitation of a fixed stepsize search in escaping local optima, we introduce a simple  mechanism called VS-RLS. In VS-RLS, the search starts with stepsize of one (the neighbourhood of a solution with one bit difference) and progressively increases the stepsize. Specifically, in each iteration, an offspring is first generated by randomly sampling from the stepsize of one of the current solution. If this solution is not nondominated to the archive, then we increase the stepsize (i.e., expand the neighbourhood of the solution), otherwise, we end this iteration and start to sample another solution in the archive for exploration again as conducted in many randomised local search methods like SEMO.
We also set a maximum stepsize (i.e., a threshold) that allows the neighbourhood to be expanded maximumly. But we do it dynamically as we want to quickly identify promising regions on which the search can focus in later stages. That is, at the beginning of the search, we set it to be the possibly maximum size (i.e., the problem size), while in the later phase of the search, when the archive gradually accumulates solutions from different promising regions, we employ a smaller maximum stepsize to focus on exploitation on those regions.

When this search mechanism is applied to the bi-objective toy example shown in Fig.~\ref{F1}, the resulting search behaviour is illustrated in Fig.~\ref{F3}. In each iteration, under a given stepsize threshold, VS-RLS first samples one candidate solution from the search region corresponding to stepsize one of the selected parent. If the sampled candidate is acceptable for archive, the archive is updated and the current iteration terminates immediately. Otherwise, the stepsize is progressively increased and a new candidate solution is sampled from the expanded search region. This process continues until an acceptable solution is found or the stepsize threshold is reached. Unlike fixed stepsize search, VS-RLS allows the exploration scale to vary during the search process rather than being confined to a single neighbourhood size. Compared with small stepsize search, the progressive expansion of the stepsize provides the likelihood of escaping from local regions and exploring more promising areas of the search space. In contrast to large fixed stepsize search, once such promising regions are reached, good solutions can often be found under the refinement of the existing ones, allowing the algorithm to stop further expansion and thus avoid wasting evaluations on unnecessarily large search regions.



As can be seen in Fig.~\ref{F3}, the fixed stepsize search is highly sensitive to the quality of the initial solution. If the starting point is close to a poor local optimum, the search often remains trapped there for a long time. Some studies have also shown that the quality of the initial solution is a major factor that affects single-solution evolutionary methods \cite{PLS2}. In contrast, VS-RLS greatly reduces its dependence on the initial solution, since it can keep expanding the search regions until a better solution is found.

\subsection{Framework of VS-RLS}
As discussed in the previous subsection, the role of the stepsize changes during the search process. A larger stepsize is mainly useful when the search needs to escape from locally optimal regions or explore distant areas of the solution space, while a smaller stepsize becomes more suitable once the search reaches potentially more promising regions. Based on this, VS-RLS can be naturally described using a two-phase strategy.


{\subsubsection{\textbf{Phase 1 (Exploration Phase)}}
At the initial phase of the search, a large stepsize threshold is adopted. Specifically, the threshold is set to the dimensionality of the decision space, allowing the stepsize to expand to its maximum possible region. In this phase, the search often starts from regions with limited improvement potential. For example, when the search is trapped in locally optimal regions or when only a small number of solutions are stored in the archive, acceptable solutions may not be reachable through small stepsize. Allowing the stepsize to expand up to the dimensionality of the decision space enables the search to explore much broader regions, thereby increasing the chance of discovering acceptable solutions that are far away from the current search location.


\subsubsection{\textbf{Phase 2 (Exploitation Phase)}}
As the search proceeds, the variable stepsize mechanism gradually guides the search toward more promising regions of the solution space. Once such regions are reached, acceptable solutions can often be obtained through small stepsize. In this phase, maintaining a large stepsize threshold becomes unnecessary and may lead to excessive evaluation cost. Therefore, the stepsize threshold is reduced to a smaller value, focusing the search on local refinement around promising regions.

Algorithm~\ref{Algorithm 2} presents the pseudocode of the proposed VS-RLS algorithm. The algorithm starts by randomly initializing a solution \(x_0\) and inserting it into the archive \(\mathcal{A}\). The search then proceeds in an iterative manner until the maximum number of function evaluations \(maxN_{fe}\) is reached.


As can be seen from the algorithm, at the beginning of each iteration, the stepsize variable $N_{cb}$ is initialized to one (Line~5). A two-phase stepsize threshold strategy is then employed. During the early phase of the search, i.e., when the iteration counter $t$ does not exceed the switching iteration $T_{vl}$, the stepsize threshold $V_L$ is set to the decision space dimension $D$ (Lines~6--7), allowing the stepsize to expand to its maximum region. After this phase, the threshold is reduced to a smaller value $V_C$ (e.g., 3) to focus the search on local refinement (Line~9). 

In each iteration, a solution is selected uniformly at random from the archive \(\mathcal{A}\) as the parent (Line~11). An offspring solution \(x'\) is then generated by randomly sampling from the neighbourhood of the parent with the current stepsize \(N_{cb}\) (Line~13). If the sampled solution \(x'\) is not dominated by any solution in the archive, it is inserted into \(\mathcal{A}\), and all archive solutions dominated by \(x'\) are removed (Lines 15--16). Otherwise, the stepsize \(N_{cb}\) is increased by one, and the sampling process is repeated (Line~19). This progressive expansion continues until either an acceptable solution is found or the stepsize threshold \(V_L\) is reached (Line~12). The algorithm then proceeds to the next iteration by updating the iteration counter and repeating the above process (Line~22).

\section{EXPERIMENTAL DESIGN}
This section first introduces the algorithms used for comparison, then presents the MOCOPs considered in this study, explains the performance indicators employed to evaluate the results, and finally describes the experimental setup in detail.

\subsection{Comparative Algorithms}
To provide a comprehensive comparison, six representative algorithms are selected, including local search methods and MOEAs. Specifically, two local search methods, three MOEAs, and one baseline method are considered. For local search methods, SEMO \cite{count35} is adopted as a representative randomized local search algorithm, which has been shown to outperform many MOEAs on MOCOPs \cite{count34}, and it is also the underlying algorithm the proposed method improves on. In addition, PLS  \cite{PLS} is included as one of the most well known Pareto-based local search techniques for MOCOPs. For MOEAs, three representative algorithms from different frameworks are selected, including NSGA-II \cite{count7}, MOEA/D \cite{count8}, and SMS-EMOA \cite{count9}. Moreover, Random Sampling (RS) \cite{count34} is considered as a baseline method. Next, we briefly introduce there algorithms.

\renewcommand{\algorithmicrequire}{\textbf{Input:}}
\renewcommand{\algorithmicensure}{\textbf{Output:}}
\begin{algorithm}[t]
\caption{VS-RLS}
\label{Algorithm 2}
\begin{algorithmic}[1]
\REQUIRE $D$: Problem size; $maxN_{fe}$: Maximum number of function evaluations; $T_{vl}$: Threshold switching iteration; $V_C$: Stepsize threshold for Phase~2. 
\ENSURE Archive $\mathcal{A}$
\STATE Randomly initialize a solution $x_0$
\STATE $\mathcal{A} \leftarrow \{x_0\}$
\STATE $t \leftarrow 1$, $N_{fe} \leftarrow 0$
\WHILE{$N_{fe} < maxN_{fe}$}
    \STATE $N_{cb} \leftarrow 1$
    \IF{$t \leq T_{vl}$}
        \STATE $V_L \leftarrow D$
    \ELSE
        \STATE $V_L \leftarrow V_C$
    \ENDIF
    \STATE Select a solution $x$ uniformly at random from $\mathcal{A}$
    \WHILE{$N_{cb} \leq V_L$}
        \STATE Randomly sample $x'$ from the neighbourhood of $x$ with scale $\leq N_{cb}$
        \STATE $N_{fe} \leftarrow N_{fe} + 1$
        \IF{$x'$ is not dominated by any solution in $\mathcal{A}$}
            \STATE $\mathcal{A} \leftarrow \mathcal{A} \cup \{x'\} \setminus \{a \in \mathcal{A} \mid x' \prec a\}$
            \STATE \textbf{break}
        \ELSE
            \STATE $N_{cb} \leftarrow N_{cb} + 1$
        \ENDIF
    \ENDWHILE
    \STATE $t \leftarrow t + 1$
\ENDWHILE
\end{algorithmic}
\end{algorithm}

\begin{enumerate}
\item SEMO \cite{count35} randomly selects a solution from the archive to perform local search. Therefore, it can be regarded as a randomized Pareto local search.
\item NSGA-II \cite{count7} is a representative Pareto-based algorithm characterized by the non-dominated sorting and crowding distance mechanisms. In NSGA-II, candidate solutions are divided into several non-dominated fronts, where solutions in lower fronts are preferred over those in higher ones. Within the same front, solutions with larger crowding distances are preferred.
\item MOEA/D \cite{count8} is a representative decomposition-based algorithm. It decomposes a multi-objective problem into multiple scalar optimization subproblems using a set of well-distributed weight vectors and a scalarizing function.
\item SMS-EMOA follows the same non-dominated sorting procedure as NSGA-II but replaces crowding distance with the hypervolume indicator to distinguish solutions within the same front. Specifically, SMS-EMOA computes the hypervolume contribution of each solution and removes the one with the smallest contribution.
\item PLS \cite{PLS} systematically explores the neighbourhood of solutions. It maintains an archive that stores all discovered non-dominated solutions. At each step, PLS randomly selects one unexplored solution from the archive and examines all of its neighbors. The archive is updated with any new non-dominated solutions found. The algorithm terminates when either the stopping criterion is satisfied or there are no unexplored solutions left in the archive.
\item RS is a simple search heuristic that randomly samples solutions from the search space. An archive is used to store all generated non-dominated solutions until the termination condition is met.
\end{enumerate}


\subsection{Multi-Objective Combinatorial Optimization Problems}
Following the practice in~\cite{count34}, four well-known MOCOPs are selected for evaluation, namely the multi-objective 0-1 knapsack \cite{KP}, traveling salesman \cite{TSP}, quadratic assignment \cite{QAP}, and NK-Landscape \cite{NK} problems. These problems are chosen because they represent diverse combinatorial structures and capture different types of search challenges. Specifically, the knapsack problem involves binary selection under capacity constraints, traveling salesman problem deals with permutation-based path optimization, quadratic assignment problem addresses complex assignment and interaction costs, and NK-Landscape problem provides tunable landscape ruggedness through the parameter \( K \).

\subsubsection{\textbf{Multi-Objective Knapsack Problem (Knapsack)}}
The Knapsack \cite{KP} involves selecting items to maximize the total value while considering multiple constraints (e.g., weight, volume) without exceeding the capacity of each knapsack. Given \( D \) items, values \( v_{ji} \) and weights \( w_{ji} \) for each item \( i \) in each knapsack \( j \), and capacity \( c_j \) for each knapsack, the problem is defined as:
\begin{equation}
\max f_j(x) = \sum_{i=1}^{D} v_{ji} \cdot x_i,\sum_{i=1}^{D} w_{ji} \cdot x_i \leq c_j, x_i \in \{0, 1\},
\end{equation}
where \( D \): Number of items; \( m \): Number of objectives; \( v_{ji} \): Value of item \( i \) for knapsack \( j \), random integers in the interval [10, 100] \cite{KP}; \( w_{ji} \): Weight of item \( i \) for knapsack \( j \), random integers in the interval [10, 100] \cite{KP}; \( c_j \): Capacity of knapsack \( j \). Set to half of the total weight of all items in the knapsack; \( x_i \): Binary variable indicating whether item \( i \) is selected.
\subsubsection{\textbf{Multi-Objective Traveling Salesman Problem (TSP)}} 
The TSP \cite{TSP} involves finding a set of Pareto optimal Hamiltonian cycles (tours) that minimize multiple travel cost objectives between cities. Given a set of cities \( V = \{v_1, v_2, \dots, v_D\} \) and a set of cost matrices \( C_j \) for \( j = 1, \dots, m \) where \( C_j(v_i, v_k) \) represents the travel cost between city \( v_i \) and city \( v_k \) under objective \( j \), the problem is defined as:
\begin{equation}
\min f_j(P) = \sum_{i=1}^{D-1} C_j(v_{i}, v_{i+1}) + C_j(v_{D}, v_{1}),
\end{equation}
where \( D \): Number of cities; \( m \): Number of objectives; \( C_j(v_i, v_k) \): Cost between city \( v_i \) and \( v_k \) for objective \( j \). This matrix is independent and generated by assigning a randomly drawn number from the interval [0,1) \cite{TSPcanshu} to each pair of cities; \( P \): A permutation representing a tour.
\subsubsection{\textbf{Multi-Objective Quadratic Assignment Problem (QAP)}} 
The QAP \cite{QAP} involves assigning a set of facilities to a set of locations such that multiple cost objectives (flow of materials, distance, etc.) between facilities are minimized. Given \( D \) facilities, \( D \) locations, cost matrices \( C_k \) for \( k = 1, \dots, m \) representing the flow between facilities, and a distance matrix \( L \) representing the distance between locations, the problem is defined as:
\begin{equation}
\min f_k(X) = \sum_{i=1}^{D} \sum_{j=1}^{D} C_k(i,j) \cdot L_{X(i)X(j)},
\end{equation}
where \(D\): Number of facilities/locations, and \(m\): Number of objectives. For each objective \(k\), the cost matrix \(C_k(i,j)\) is independently generated, with every entry uniformly drawn from the range \([0,100]\) \cite{QAP}. To construct the distance matrix, \(D\) locations are randomly placed on a two-dimensional plane within the range \([0,5000]\) \cite{QAPcanshu}, and the Euclidean distance between each pair of locations \(u\) and \(v\) defines \(L_{uv}\). 
\subsubsection{\textbf{Multi-Objective NK-Landscape Problem}}
The NK-Landscape \cite{NK} is a problem where each bit in a binary string interacts with \( K \) other bits, and the objective is to maximize multiple fitness functions, which depend on these interactions. Given a binary string \( x \) of stepsize \( D \), and each bit \( x_i \) has an associated contribution function \( c_{ij} \) depending on \( K \) interacting bits, the problem is defined as:
\begin{equation}
\max f_j(x) = \frac{1}{D} \sum_{i=1}^{D} c_{ij}(x_i, x_{k_{i1}}, \dots, x_{k_{iK}}),
\end{equation}
where \( D \): size of the binary string; \( m \): Number of objectives; \( K \): Number of interacting bits. This paper K=10 \cite{count34}; \( c_{ij} \): Contribution function for bit \( x_i \) in objective \( j \); \( x_{k_{i1}}, \dots, x_{k_{iK}} \): Interacting bits for bit \( x_i \).

\subsection{Performance Indicator}
The hypervolume (HV) indicator~\cite{KP} is used to assess the performance of multi-objective optimization algorithms. It measures the volume of the objective space dominated by the obtained solution set with respect to a reference point. In the absence of the true Pareto front, HV is considered one of the most effective performance indicators \cite{li2019quality}. When calculating HV, the choice of the reference point is critical. It is typically set slightly worse than the worst objective values to avoid a zero HV. However, this may still result in zero HV values for poorly performing algorithms, which is common in MOCOPs. To address this issue, following the practice in \cite{count34}, we use random sampling as a baseline to determine the position of the reference point. The reference point \( r \) is set slightly worse than the nadir point. According to~\cite{50}, for a minimization problem, the reference value for the \( i \)-th objective is given by $r_i = \text{max}_i + (\text{max}_i - \text{min}_i)/{10}$, where \( \text{max}_i \) and \( \text{min}_i \) denote the maximum and minimum values of the non-dominated set (obtained by random sampling) for the \( i \)-th objective, respectively. For a maximization problem, the reference point is set in the opposite direction as $r_i = \text{min}_i - (\text{max}_i - \text{min}_i)/{10}$. 

\begin{table*}[t]
\centering
\caption{The HV results (mean and standard deviation) of the seven algorithms under 4 different settings. The best-performing algorithm in each setting is highlighted in bold. The symbols ``+", ``=", and ``$-$" indicate that the corresponding algorithm performs statistically better than, similar to, or worse than VS-RLS, respectively.}
	\resizebox{\textwidth}{!}{
		\begin{tabular}{@{}ccccccccc@{}}
			\toprule
			Problem    & D     & RS & NSGA-II & MOEA/D & SMS-EMOA & PLS   & SEMO  & VS-RLS \\
			\midrule
			Knapsack   & 500   & 3.38e+07 (6.85e+05)$^{-}$ & 9.02e+07 (7.28e+05)$^{-}$ & 9.07e+07 (4.94e+05)$^{-}$ & 8.82e+07 (4.92e+05)$^{-}$ & 2.96e+07 (4.82e+06)$^{-}$ & 9.04e+07 (9.93e+05)$^{-}$ & \textbf{9.18e+07 (5.59e+05)} \\
			TSP        & 200   & 1.54e+03 (3.60e+01)$^{-}$ & 9.00e+03 (1.09e+02)$^{-}$ & 9.74e+03 (1.22e+02)$^{-}$ & 8.82e+03 (1.36e+02)$^{-}$ & 1.64e+03 (1.30e+03)$^{-}$ & 9.82e+03 (8.09e+01)$^{-}$ & \textbf{9.98e+03 (7.99e+01)} \\
			QAP        & 100   & 6.42e+14 (2.28e+13)$^{-}$ & 2.61e+15 (1.62e+14)$^{-}$ & 2.56e+15 (1.20e+14)$^{-}$ & 2.60e+15 (1.30e+14)$^{-}$ & 2.57e+15 (1.21e+15)$^{-}$ & 6.14e+15 (1.20e+14)$^{-}$ & \textbf{6.21e+15 (1.10e+14)} \\
			NK-landscape        & 100   & 8.01e--02 (2.10e--03)$^{-}$ & 1.44e--01 (5.22e--03)$^{-}$ & 1.40e--01 (6.18e--03)$^{-}$ & 1.45e--01 (5.08e--03)$^=$ & 1.04e--01 (2.41e--02)$^{-}$ & 1.39e--01 (4.29e--03)$^{-}$ & \textbf{1.47e--01 (4.62e--03)} \\
			\midrule
			\multicolumn{2}{c}{+/=/$-$} & 0/0/4 & 0/0/4 & 0/0/4 & 0/1/3 & 0/0/4 & 0/0/4 &  \\
			\bottomrule
	\end{tabular}}%
	\label{Table1}%
\end{table*}%

\begin{figure*}[t]
	\centering  
	\subfigure[Knapsack]{
		\includegraphics[width=1.65in]{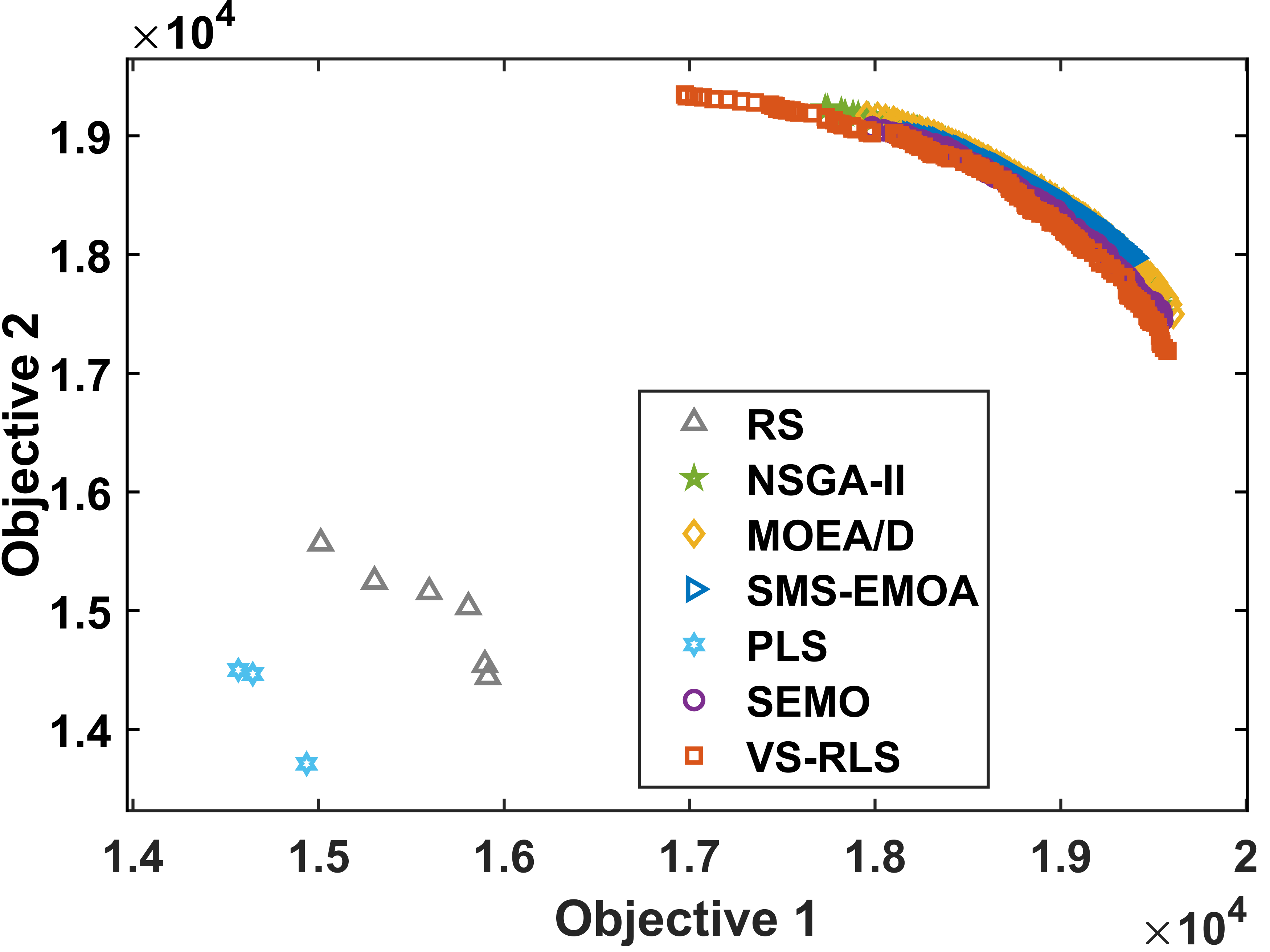} }
	\subfigure[TSP]{
		\includegraphics[width=1.65in]{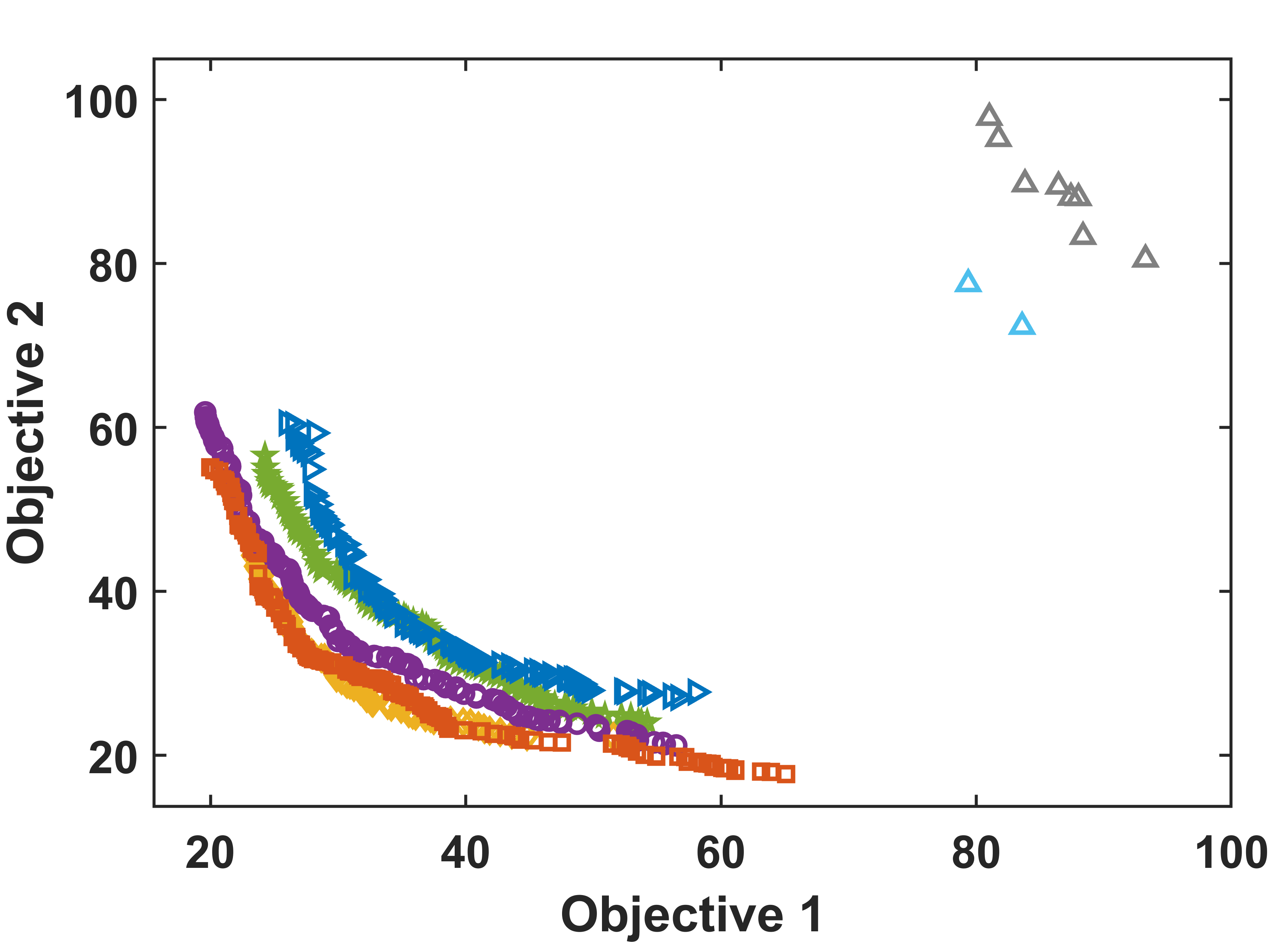} }
	\subfigure[QAP]{
		\includegraphics[width=1.65in]{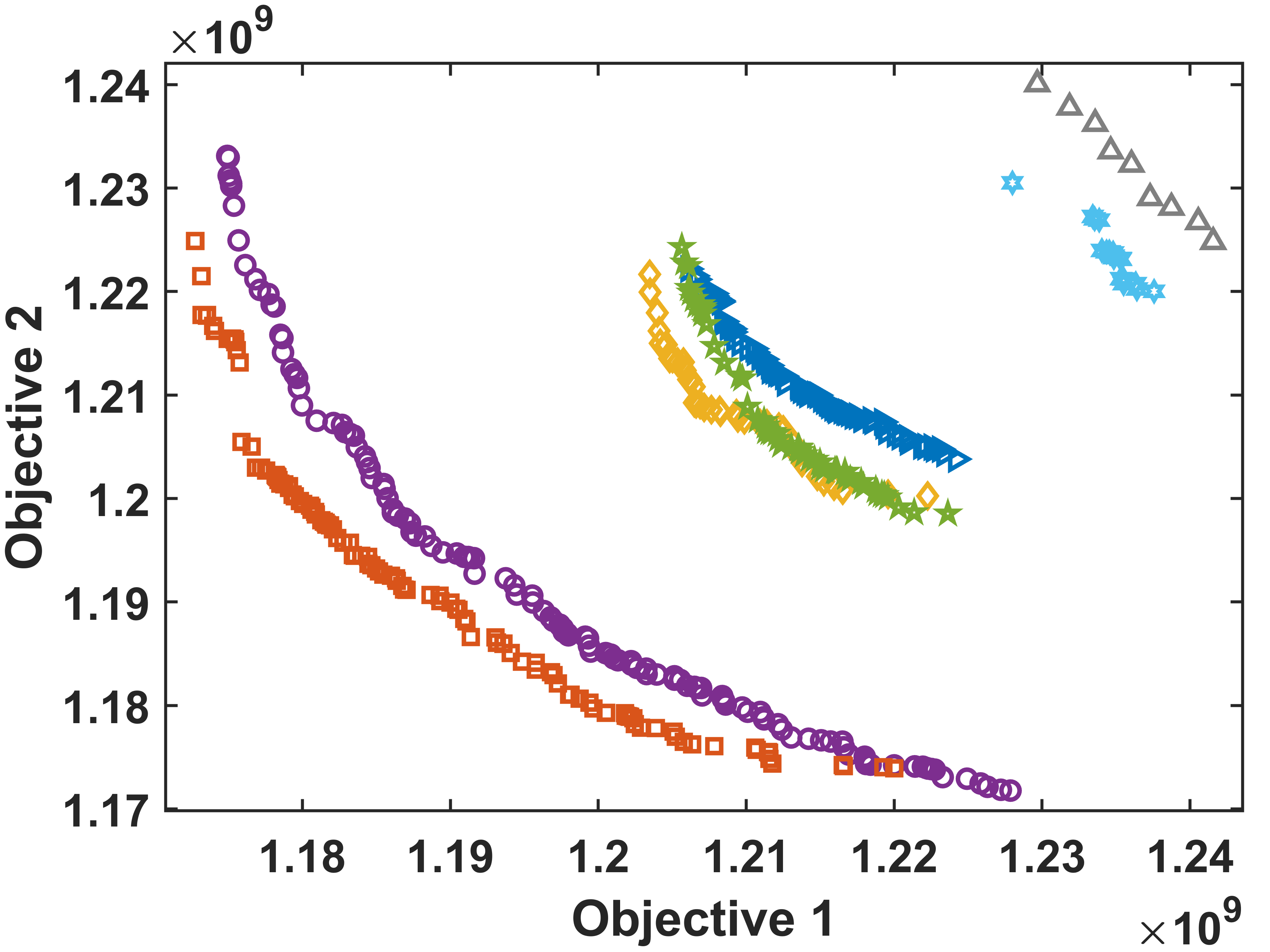} }
	\subfigure[NK-landscape]{
		\includegraphics[width=1.65in]{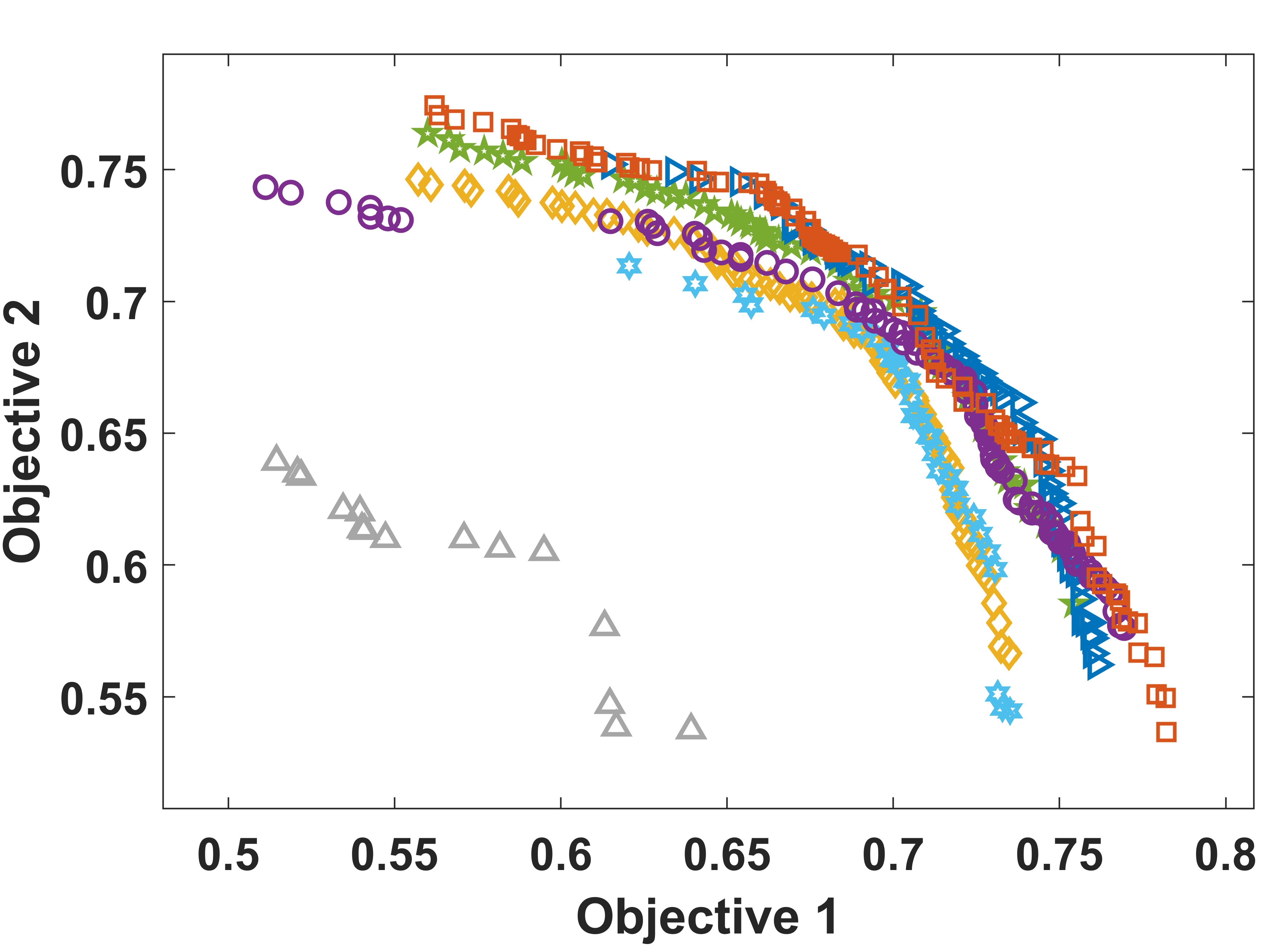} }
	\caption{The final archive sets are the results of the seven algorithms on 4 different settings of the first experiment from a single run. This particular run is associated with the result closest to the average HV value.}
	\label{Figure9}
\end{figure*}

\subsection{Experimental Setup}
In general, local search algorithms require a substantial search budget~\cite{count34}. When the budget is severely limited, local search can explore only a small number of solutions. In contrast, MOEAs typically need fewer evaluations and can evolve more quickly, though they may still become trapped in local optima even with sufficient resources~\cite{count6}. To comprehensively evaluate the performance of the algorithms, we conducted three sets of experiments. The parameters were set as $T_{vl}=1000$ and $V_C=3$. The population size for the MOEAs is set to 100, and each algorithm is run independently 30 times.

We first tested the proposed algorithm on four representative problems, the multi-objective Knapsack, TSP, QAP, and NK-Landscape, with respective sizes of \(D=500\), \(200\), \(100\), and \(100\), respectively. Each run was allowed up to \(10^6\) evaluations. 


Next, we examined the scalability of the algorithms on different problem sizes~\cite{count34}. Based on the first set of experiments, we further tested the Knapsack problem with \(D=100\) and \(1000\), the TSP with \(D=50\) and \(500\), and both the QAP and NK-Landscape problems with \(D=50\) and \(200\). 


Finally, we investigated the influence of different number of evaluations~\cite{MOEADTSP}. Using the same problem settings as in the first experiment, we conducted tests with total evaluations of \(1\times10^5\), \(5\times10^5\), \(2\times10^6\), and \(5\times10^6\).


Additionally, appropriate search operators need to be set for the algorithms. 
\subsubsection{\textbf{For MOEAs}} The Knapsack and NK-landscape problems utilize uniform crossover (crossover rate of 1.0) \cite{canshu} and bit-flip mutation (mutation rate of $1/D$ \cite{canshu}, where $D$ represents the number of variables). The TSP problem is based on sequence permutation, whereas the QAP problem is based on position permutation. Therefore, the TSP problem employs order crossover and 2-opt mutation \cite{canshu}, which are suited for sequence permutations, while the QAP problem employs cycle crossover and 2-swap mutation \cite{canshu}, which are suited for position permutation. For these two types of permutation problems, the crossover rate is set to 1.0, and the mutation rate is set to $0.05$ \cite{count34}. 
\subsubsection{\textbf{For Local Search}}

Bit flip operators are adopted with different neighbourhood definitions: a 2-bit flip neighbourhood for the Knapsack problem and a 1-bit flip neighbourhood for the NK-landscape problem. For the Knapsack problem, 1-bit flip can lead to either improvement on both objectives or degradation on them according to the problem definition, which is not reasonable and helpful for local search. Therefore, a 2-bit flip neighbourhood is adopted for Knapsack \cite{count34}}. For permutation-based problems, 2-opt operators are used for the TSP, and 2-swap operators are used for the QAP.


\section{PERFORMANCE VERIFICATION OF VS-RLS}
In this section, we first compare the performance of VS-RLS under general settings. Then, we examine its behavior with different problem sizes. Next, we analyze the effect of different maximum function evaluations on the results. Finally, we study the sensitivity of the parameters and perform an ablation study to understand the role of each component in VS-RLS.

\subsection{Comparison at general settings}
Table~\ref{Table1} presents the mean and standard deviation of the HV values obtained by the seven algorithms on the four test instances. 
For each setting, the best-performing algorithm is highlighted in bold. 
The symbols “+”, “=”, and “–” indicate that the corresponding algorithm performs significantly better than, equivalent to, or worse than VS-RLS, respectively, according to the Wilcoxon rank-sum test at a significance level of 0.05. Fig.~\ref{Figure9} shows the final solution set from a single run whose result is closest to the average HV value for this problem. The same rule applies to the tables and figures presented in the following experimental results.


From Table~\ref{Table1} and Fig.~\ref{Figure9}, it can be seen that the proposed VS-RLS algorithm achieves the best overall performance across all four types of test instances. In the Knapsack problem (Fig.~\ref{Figure9}(a)), MOEAs usually show slightly better convergence, while VS-RLS demonstrates a stronger ability to maintain solution diversity. Specifically, VS-RLS performs a single-solution local search guided by a variable stepsize mechanism, which can flexibly adjust the search region and explore different regions of the solution space. This adaptive mechanism helps VS-RLS maintain diverse non-dominated solutions even when its convergence speed is relatively slower. A more detailed discussion of this behavior under different problem sizes will be presented later on.

In the TSP problem (Fig.~\ref{Figure9}(b)), NSGA-II and SMS-EMOA perform similarly, whereas MOEA/D shows slightly better convergence. A possible explanation is that TSP often contains multiple large funnel structures with many small attraction basins~\cite{TSP1}, where algorithms may converge slowly toward these local basins~\cite{TSP2}. MOEA/D decomposes the problem using uniformly distributed weight vectors, whose search directions may align with certain attraction basins in TSP, enabling it to move more smoothly toward global regions. Although MOEA/D obtains a slightly lower HV value than VS-RLS, this property may provide a potential advantage in high-dimensional TSP problems~\cite{MOEADTSP}, as will be further discussed in later experiments.

For the QAP (Fig.~\ref{Figure9}(c)) and NK-landscape (Fig.~\ref{Figure9}(d)) problems, VS-RLS shows the best performance. In QAP, population-based MOEAs often stop improving too early near the Pareto front. This may be because the objective values in QAP cover a very wide range, which increases selection pressure and causes the population to converge quickly to local optima. In contrast, the NK-landscape has smaller changes in objective values, which result in weaker selection pressure and slower convergence. VS-RLS works well in both situations by changing its search range dynamically. It enlarges the range to escape stagnation in QAP and reduces it to make steady improvements in the NK-landscape. This makes VS-RLS robust across problems with different landscape features.

\begin{table*}[!t]
\centering
\caption{The HV results (mean and standard deviation) of the seven algorithms under 8 different settings. The best-performing algorithm in each setting is highlighted in bold. The symbols ``+", ``=", and ``$-$" indicate that the corresponding algorithm performs statistically better than, similar to, or worse than VS-RLS, respectively.}
	\resizebox{\textwidth}{!}{
    \begin{tabular}{@{}ccccccccc@{}}
		\toprule
		Problem    & D     & RS & NSGA-II & MOEA/D & SMS-EMOA & PLS   & SEMO  & VS-RLS \\
		\midrule
		\multirow{2}[1]{*}{Knapsack}  & 100   & 5.88e+06 (1.14e+05)$^{-}$ & 8.31e+06 (3.74e+04)$^{=}$ & 8.29e+06 (3.88e+04)$^{-}$ & 8.23e+06 (4.13e+04)$^{-}$ & 6.84e+06 (2.40e+05)$^{-}$ & 8.28e+06 (3.40e+04)$^{-}$ & \textbf{8.32e+06 (2.90e+04)} \\
	  & 1000  & 7.33e+07 (1.57e+06)$^{-}$ & 2.67e+08 (1.81e+06)$^{-}$ & 
        2.72e+08 (1.50e+06)$^{-}$ & 2.62e+08 (1.36e+06)$^{-}$ & 5.30e+07 (5.02e+06)$^{-}$ & 2.70e+08 (1.86e+06)$^{-}$ & \textbf{2.75e+08 (1.39e+06)} \\
		\multirow{2}[0]{*}{TSP}      & 50    & 3.57e+02 (5.53e+00)$^{-}$ & 8.86e+02 (1.11e+01)$^{-}$ & 8.87e+02 (1.32e+01)$^{-}$ & 8.76e+02 (1.16e+01)$^{-}$ & 6.12e+02 (6.79e+01)$^{-}$ & 9.36e+02 (5.27e+00)$^{-}$ & \textbf{9.42e+02 (4.84e+00)} \\
		& 500   & 3.97e+03 (7.53e+01)$^{-}$ & 3.85e+04 (1.02e+03)$^{-}$ & \textbf{4.50e+04 (6.64e+02)$^{+}$} & 3.78e+04 (6.84e+02)$^{-}$ & 2.49e+03 (2.19e+02)$^{-}$ & 4.20e+04 (3.55e+02)$^{-}$ & 4.30e+04 (4.27e+02) \\
		\multirow{2}[0]{*}{QAP}        & 50    & 3.13e+14 (6.21e+12)$^{-}$ & 5.43e+14 (2.19e+13)$^{-}$ & 5.22e+14 (2.13e+13)$^{-}$ & 5.38e+14 (2.62e+13)$^{-}$ & 4.71e+14 (2.42e+13)$^{-}$ & 7.95e+14 (1.56e+13)$^{-}$ & \textbf{8.03e+14 (1.67e+13)} \\
		& 200   & 4.60e+15 (1.20e+14)$^{-}$ & 1.52e+16 (5.89e+14)$^{-}$ & 1.53e+16 (7.70e+14)$^{-}$ & 1.46e+16 (6.40e+14)$^{-}$ & 8.49e+15 (5.11e+15)$^{-}$ & 3.04e+16 (5.95e+14)$^{-}$ & \textbf{3.08e+16 (4.86e+14)} \\
		\multirow{2}[1]{*}{NK-landscape}   & 50    & 1.50e--01 (2.99e--03)$^{-}$ & 1.91e--01 (8.42e--03)$^{=}$ & 1.88e--01 (6.78e--03)$^{-}$ & \textbf{1.96e--01 (6.81e--03)$^{=}$} & 1.51e--01 (2.12e--02)$^{-}$ & 1.78e--01 (9.98e--03)$^{-}$ & 1.95e--01 (4.87e--03) \\
		& 200   & 3.67e--02 (1.20e--03)$^{-}$ & 1.01e--01 (3.95e--03)$^{-}$ & 1.02e--01 (3.17e--03)$^{-}$ & 1.01e--01 (3.25e--03)$^{-}$ & 8.18e--02 (2.15e--02)$^{-}$ & 1.05e--01 (2.47e--03)$^{=}$ & \textbf{1.06e--01 (2.69e--03)} \\
		\midrule
		\multicolumn{2}{c}{+/=/$-$} & 0/0/8 & 0/2/6 & 1/0/7 & 0/1/7 & 0/0/8 & 0/1/7 &  \\
		\bottomrule
	\end{tabular}}
	\label{Table2}
\end{table*}

\begin{figure*}[t]
	\centering  
	\subfigure[Knapsack on D=100]{
		\includegraphics[width=1.6in]{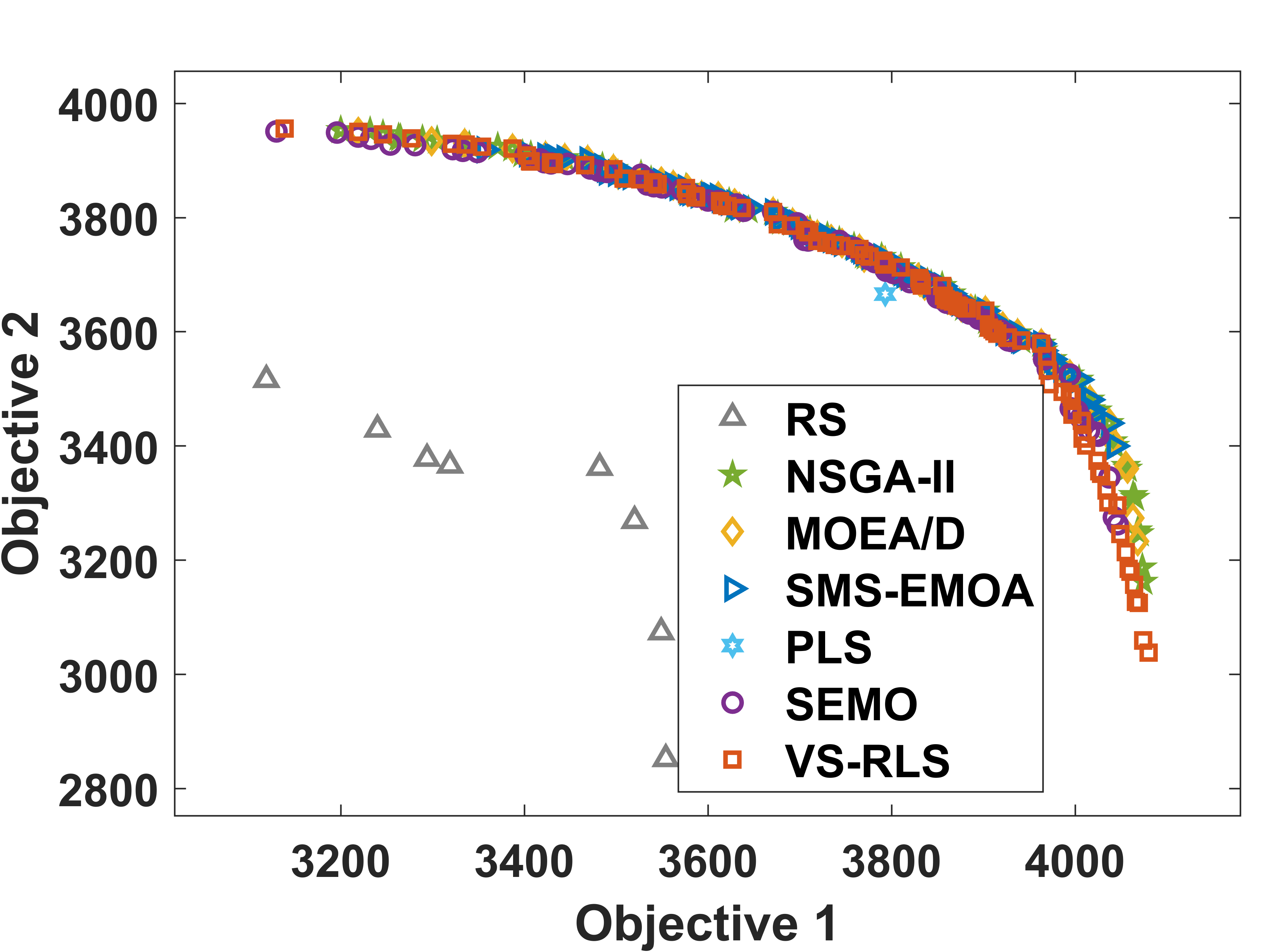} }
	\subfigure[Knapsack on D=1000]{
		\includegraphics[width=1.6in]{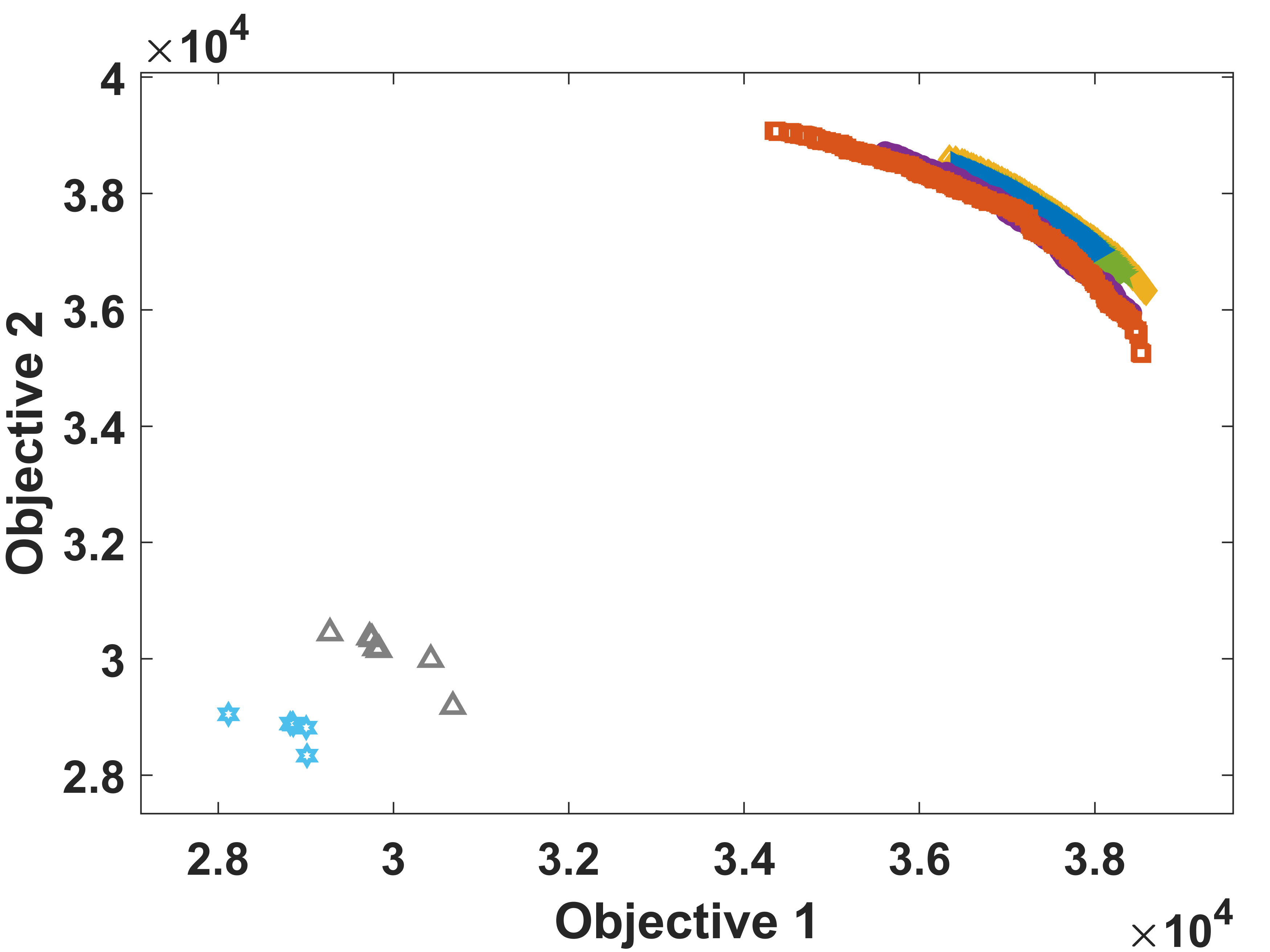} }
	\subfigure[TSP on D=500]{
		\includegraphics[width=1.6in]{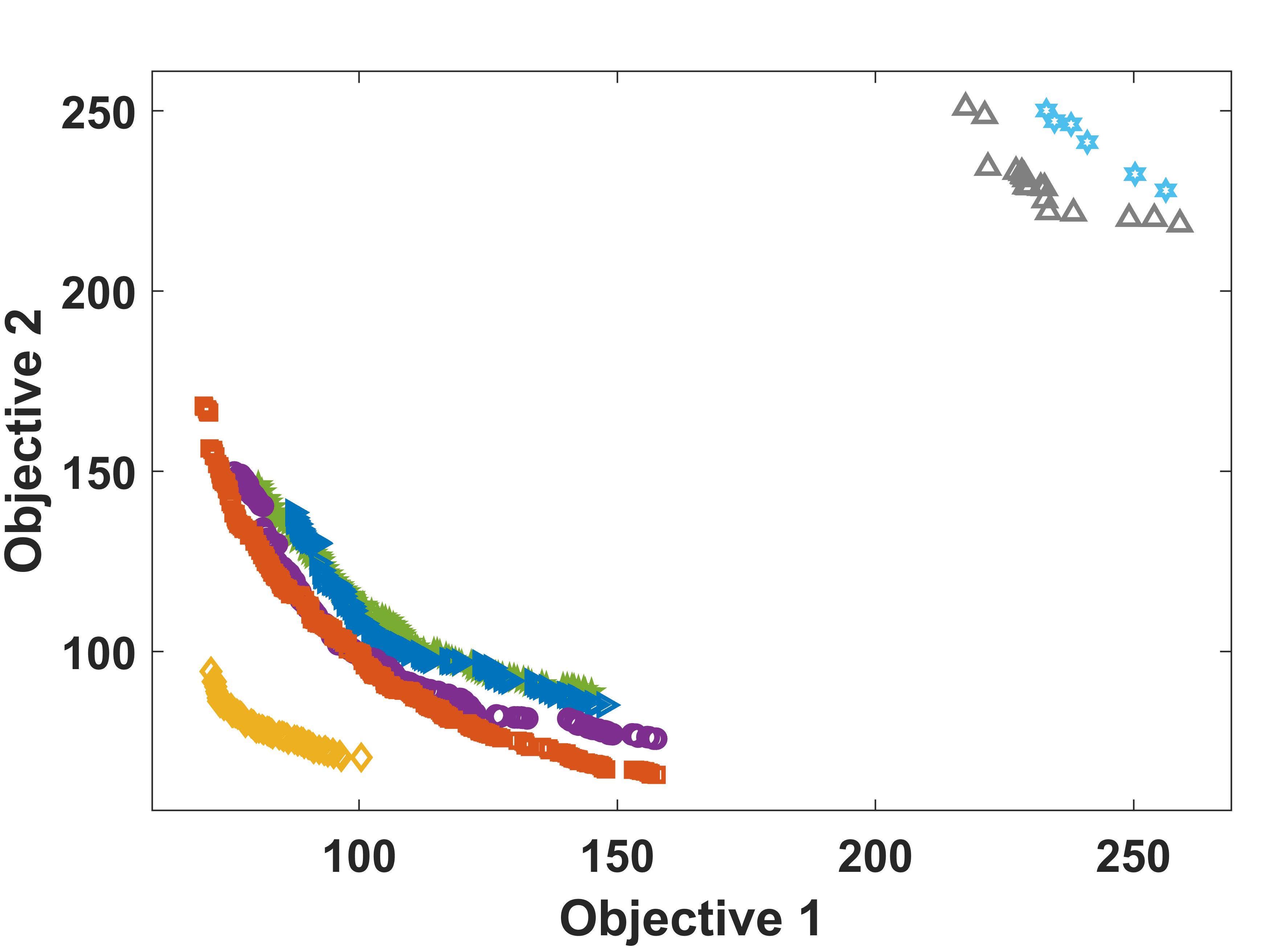} }
	\subfigure[NK-landscape on D=50]{
		\includegraphics[width=1.6in]{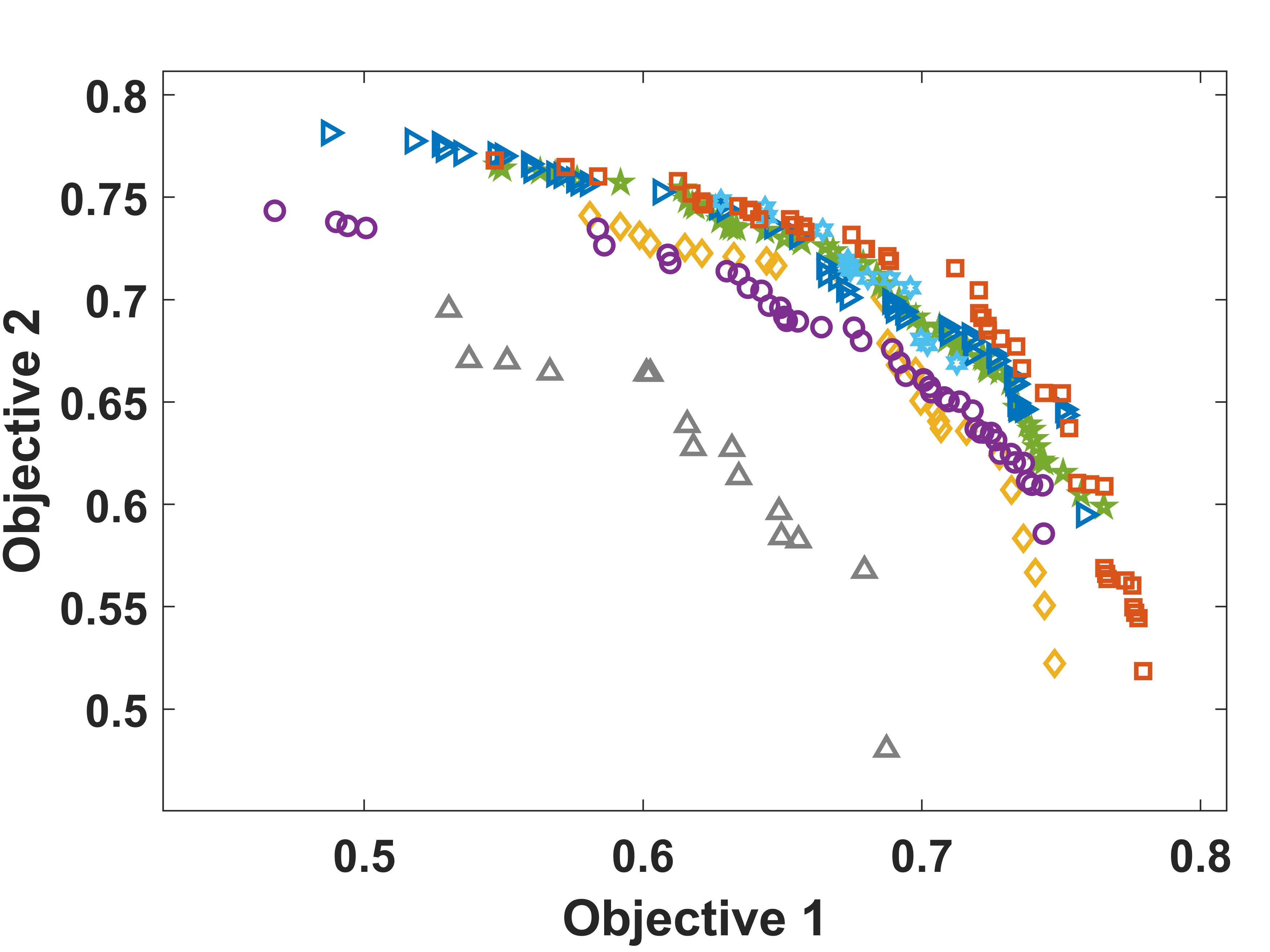} }
	\caption{The final archive sets are the results of the seven algorithms on 4 different settings of the second experiment from a single run. This particular run is associated with the result closest to the average HV value.}
	\label{Figure10}
\end{figure*}

\subsection{Comparison at different problem sizes}
Table~\ref{Table2} presents the mean and standard deviation of the HV values obtained by the seven algorithms on the 8 test instances. Fig.~\ref{Figure10} shows the final solution set from a single run whose result is closest to the average HV value for this problem.


This experiment studies the performance of the algorithms under different problem sizes. As shown in Table~\ref{Table2} and Fig.~\ref{Figure10}, the proposed VS-RLS achieves the best performance in most instances. For the Knapsack problem (Figs.~\ref{Figure10}(a) and (b)), when the dimension is low (\(D = 100\)), the differences among the algorithms are small and most of them can approximate the Pareto front well. As the dimension increases, MOEAs tend to produce solutions with stronger convergence but lower diversity, while VS-RLS maintains better diversity and correspondingly achieves higher HV values. A possible explanation is that, in high-dimensional Knapsack problems, the search process is more likely to concentrate around certain local Pareto fronts, making it difficult for population-based algorithms with fixed crossover and mutation operators to preserve sufficient diversity. In contrast, VS-RLS performs local search on a single solution with a variable stepsize mechanism, which allows it to adaptively adjust the search region and escape from locally concentrated regions. Although its convergence may be slightly slower in some cases, the final solution set obtained by VS-RLS demonstrates better diversity and coverage of the Pareto front.

In the TSP problem (Fig.~\ref{Figure10}(c)), compared with the first experiment, the most notable change is the clear improvement in MOEA/D’s performance, which becomes better as the dimension increases~\cite{MOEADTSP}.

In the NK-landscape problem (Fig.~\ref{Figure10}(d)), as the dimension \( D \) increases, the relative ruggedness level of the landscape decreases. In small-scale instances (such as \( D \)=50), the performance of SMS-EMOA is similar to that of VS-RLS (Table \ref{Table2}), with a slightly higher HV value. 
In high-dimensional instances, VS-RLS can better balance exploration and exploitation through its dynamic stepsize adjustment, thus achieving better results.

\begin{table*}[t]
\centering
\caption{The HV results (mean and standard deviation) of the seven algorithms under 16 different settings. The best-performing algorithm in each setting is highlighted in bold. The symbols ``+", ``=", and ``$-$" indicate that the corresponding algorithm performs statistically better than, similar to, or worse than VS-RLS, respectively.}
	\resizebox{\textwidth}{!}{
    \begin{tabular}{@{}cccccccccc@{}}
			\toprule
			Problem & eval.   & D     & RS & NSGA-II & MOEA/D & SMS-EMOA & PLS   & SEMO  & VS-RLS \\
			\midrule
			\multirow{4}[1]{*}{Knapsack} & \(1\times10^5\) & \multirow{4}[1]{*}{500} & 3.15e+07 (7.19e+05)$^{-}$ & 8.45e+07 (6.34e+05)$^{+}$ & 8.46e+07 (8.26e+05)$^{+}$ & 8.44e+07 (6.07e+05)$^{+}$ & 2.40e+07 (1.16e+06)$^{-}$ & \textbf{8.46e+07 (8.08e+05)$^{+}$} & 7.06e+07 (3.66e+06) \\
			& \(5\times10^5\) &       & 3.31e+07 (9.03e+05)$^{-}$ & 8.83e+07 (6.10e+05)$^{-}$ & 8.93e+07 (5.44e+05)$^{-}$ & 8.70e+07 (4.97e+05)$^{-}$ & 2.68e+07 (3.15e+06)$^{-}$ & 8.88e+07 (5.72e+05)$^{-}$ & \textbf{8.99e+07 (4.25e+05)} \\
			& \(2\times10^6\) &       & 3.45e+07 (8.68e+05)$^{-}$ & 9.19e+07 (5.23e+05)$^{-}$ & 9.22e+07 (5.40e+05)$^{-}$ & 8.92e+07 (5.62e+05)$^{-}$ & 3.81e+07 (7.77e+06)$^{-}$ & 9.21e+07 (5.92e+05)$^{-}$ & \textbf{9.35e+07 (5.36e+05)} \\
			& \(5\times10^6\) &       & 3.53e+07 (8.03e+05)$^{-}$ & 9.41e+07 (5.39e+05)$^{-}$ & 9.38e+07 (3.03e+05)$^{-}$ & 9.10e+07 (5.46e+05)$^{-}$ & 5.71e+07 (1.33e+07)$^{-}$ & 9.39e+07 (4.71e+05)$^{-}$ & \textbf{9.52e+07 (4.24e+05)} \\
			\multirow{4}[0]{*}{TSP} & \(1\times10^5\) & \multirow{4}[0]{*}{200} & 1.40e+03 (4.46e+01)$^{-}$ & 6.82e+03 (2.74e+02)$^{-}$ & \textbf{7.84e+03 (1.69e+02)$^{+}$} & 6.58e+03 (1.93e+02)$^{-}$ & 1.04e+03 (1.40e+02)$^{-}$ & 7.28e+03 (1.51e+02)$^{-}$ & 7.44e+03 (1.14e+02) \\
			& \(5\times10^5\) &       & 1.51e+03 (3.78e+01)$^{-}$ & 8.37e+03 (1.97e+02)$^{-}$ & 9.25e+03 (1.43e+02)$^{=}$ & 8.24e+03 (1.99e+02)$^{-}$ & 1.27e+03 (5.53e+02)$^{-}$ & 9.14e+03 (7.49e+01)$^{-}$ & \textbf{9.23e+03 (9.37e+01)} \\
			& \(2\times10^6\) &       & 1.58e+03 (3.00e+01)$^{-}$ & 9.50e+03 (1.50e+02)$^{-}$ & 1.01e+04 (1.24e+02)$^{-}$ & 9.35e+03 (1.50e+02)$^{-}$ & 3.10e+03 (3.14e+03)$^{-}$ & 1.04e+04 (6.08e+01)$^{-}$ & \textbf{1.06e+04 (5.39e+01)} \\
			& \(5\times10^6\) &       & 1.64e+03 (2.80e+01)$^{-}$ & 1.01e+04 (9.94e+01)$^{-}$ & 1.05e+04 (1.19e+02)$^{-}$ & 1.00e+04 (1.32e+02)$^{-}$ & 4.84e+03 (4.02e+03)$^{-}$ & 1.11e+04 (5.02e+01)$^{-}$ & \textbf{1.12e+04 (6.10e+01)} \\
			\multirow{4}[0]{*}{QAP} & \(1\times10^5\) & \multirow{4}[0]{*}{100} & 5.49e+14 (2.93e+13)$^{-}$ & 1.91e+15 (1.77e+14)$^{-}$ & 2.00e+15 (1.16e+14)$^{-}$ & 1.83e+15 (1.51e+14)$^{-}$ & 8.95e+14 (6.19e+14)$^{-}$ & \textbf{4.25e+15 (1.20e+14)$^{=}$} & 4.23e+15 (1.21e+14) \\
			& \(5\times10^5\) &       & 6.10e+14 (2.61e+13)$^{-}$ & 2.41e+15 (1.47e+14)$^{-}$ & 2.45e+15 (1.07e+14)$^{-}$ & 2.49e+15 (1.53e+14)$^{-}$ & 2.70e+15 (1.14e+15)$^{-}$ & 5.61e+15 (1.31e+14)$^{=}$ & \textbf{5.62e+15 (1.25e+14)} \\
			& \(2\times10^6\) &       & 6.73e+14 (2.55e+13)$^{-}$ & 2.77e+15 (1.65e+14)$^{-}$ & 2.70e+15 (1.90e+14)$^{-}$ & 2.81e+15 (1.53e+14)$^{-}$ & 3.23e+15 (5.17e+14)$^{-}$ & 6.59e+15 (1.47e+14)$^{-}$ & \textbf{6.68e+15 (1.20e+14)} \\
			& \(5\times10^6\) &       & 7.02e+14 (2.86e+13)$^{-}$ & 2.93e+15 (1.61e+14)$^{-}$ & 2.91e+15 (1.72e+14)$^{-}$ & 2.97e+15 (1.11e+14)$^{-}$ & 3.35e+15 (2.70e+14)$^{-}$ & 7.09e+15 (8.63e+13)$^{-}$ & \textbf{7.18e+15 (1.38e+14)} \\
			\multirow{4}[1]{*}{NK-landscape} & \(1\times10^5\) & \multirow{4}[1]{*}{100} & 7.36e--02 (2.46e--03)$^{-}$ & 1.34e--01 (3.19e--03)$^{=}$ & 1.37e--01 (6.59e--03)$^{=}$ & 1.34e--01 (5.27e--03)$^{=}$ & 1.09e--01 (2.37e--02)$^{-}$ & \textbf{1.42e--01 (5.44e--03)$^{+}$} & 1.36e--01 (4.81e--03) \\
			& \(5\times10^5\) &       & 7.80e--02 (1.75e--03)$^{-}$ & 1.41e--01 (3.93e--03)$^{-}$ & 1.42e--01 (5.52e--03)$^{-}$ & 1.43e--01 (4.73e--03)$^{=}$ & 1.14e--01 (2.06e--02)$^{-}$ & 1.40e--01 (4.36e--03)$^{-}$ & \textbf{1.45e--01 (4.74e--03)} \\
			& \(2\times10^6\) &       & 8.21e--02 (2.06e--03)$^{-}$ & 1.45e--01 (6.08e--03)$^{-}$ & 1.42e--01 (4.60e--03)$^{-}$ & 1.45e--01 (4.52e--03)$^{-}$ & 1.12e--01 (2.21e--02)$^{-}$ & 1.43e--01 (5.02e--03)$^{-}$ & \textbf{1.50e--01 (5.09e--03)} \\
			& \(5\times10^6\) &       & 8.51e--02 (1.87e--03)$^{-}$ & 1.46e--01 (4.67e--03)$^{-}$ & 1.45e--01 (5.68e--03)$^{-}$ & 1.47e--01 (4.00e--03)$^{-}$ & 1.14e--01 (2.25e--02)$^{-}$ & 1.41e--01 (5.92e--03)$^{-}$ & \textbf{1.50e--01 (4.54e--03)} \\
			\midrule
			\multicolumn{3}{c}{+/=/$-$} & 0/0/16 & 1/1/14 & 2/2/12 & 1/2/13 & 0/0/16 & 2/1/13 &  \\
			\bottomrule
	\end{tabular}}%
	\label{Table3}%
\end{table*}%

\begin{figure*}[!t]
	\centering  
	\subfigure[Knapsack on eval.=\(1\times10^5\)]{
		\includegraphics[width=1.6in]{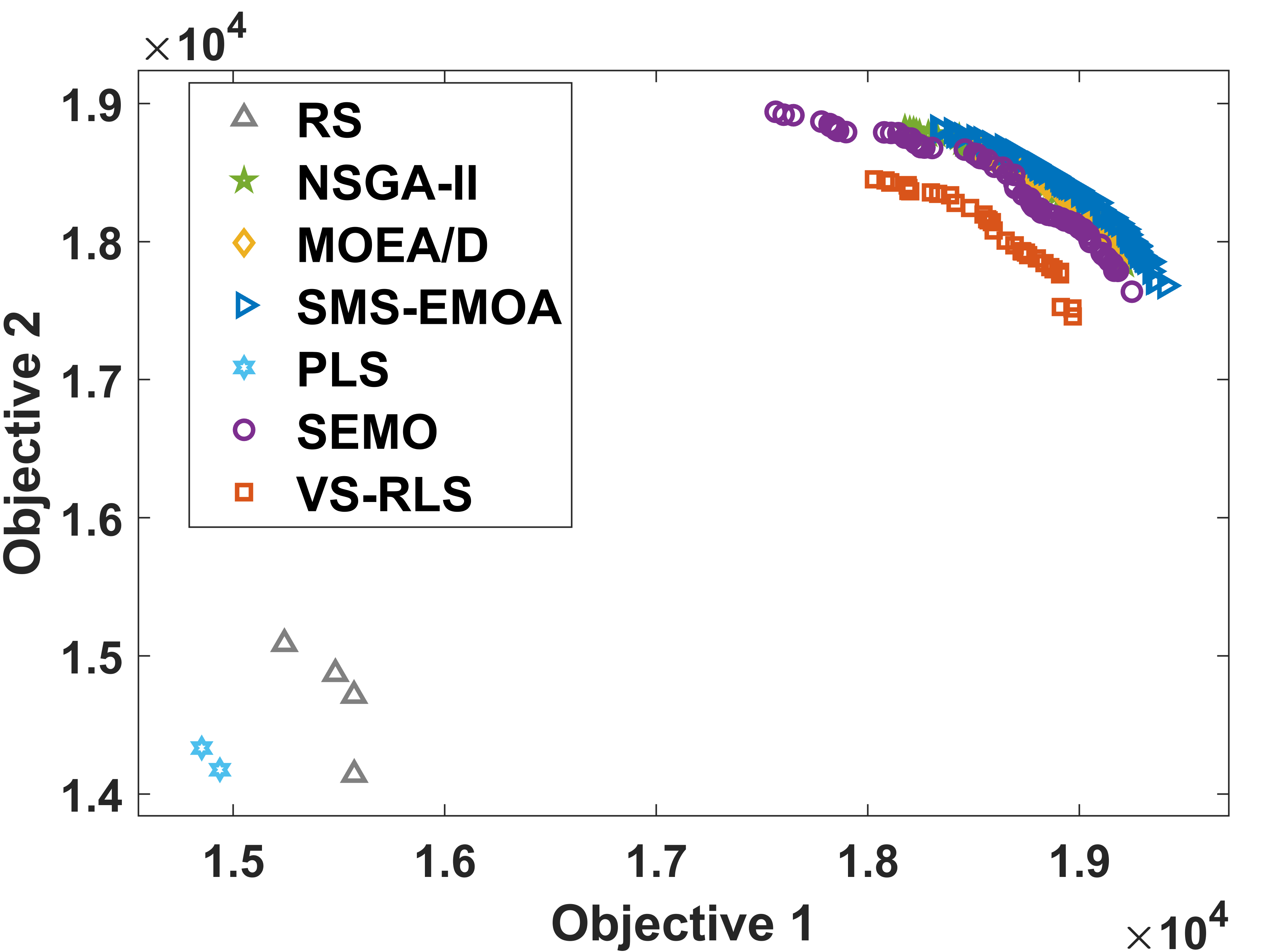} }
	\subfigure[Knapsack on eval.=\(5\times10^5\)]{
		\includegraphics[width=1.6in]{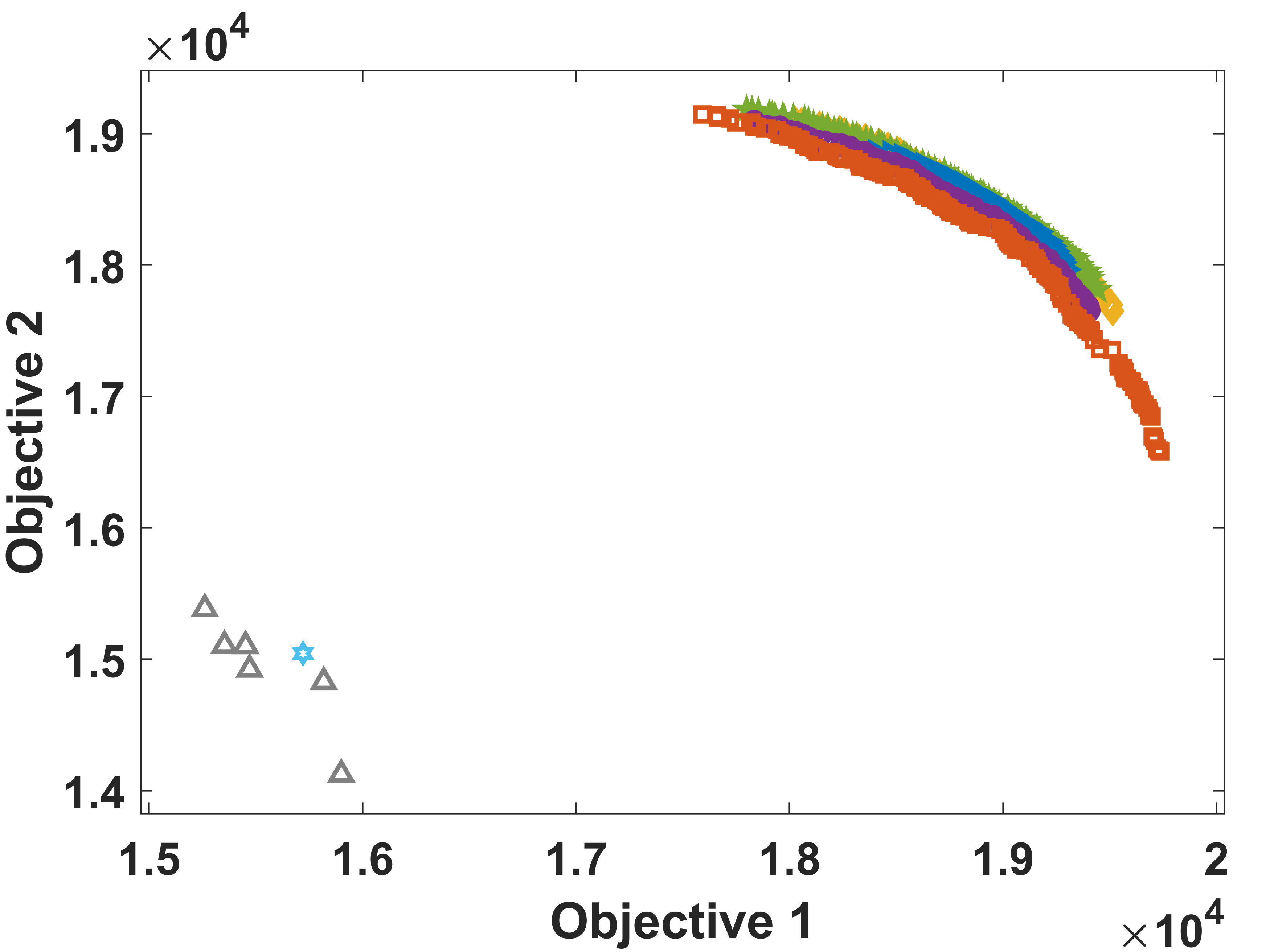} }
	\subfigure[Knapsack on eval.=\(2\times10^6\)]{
		\includegraphics[width=1.6in]{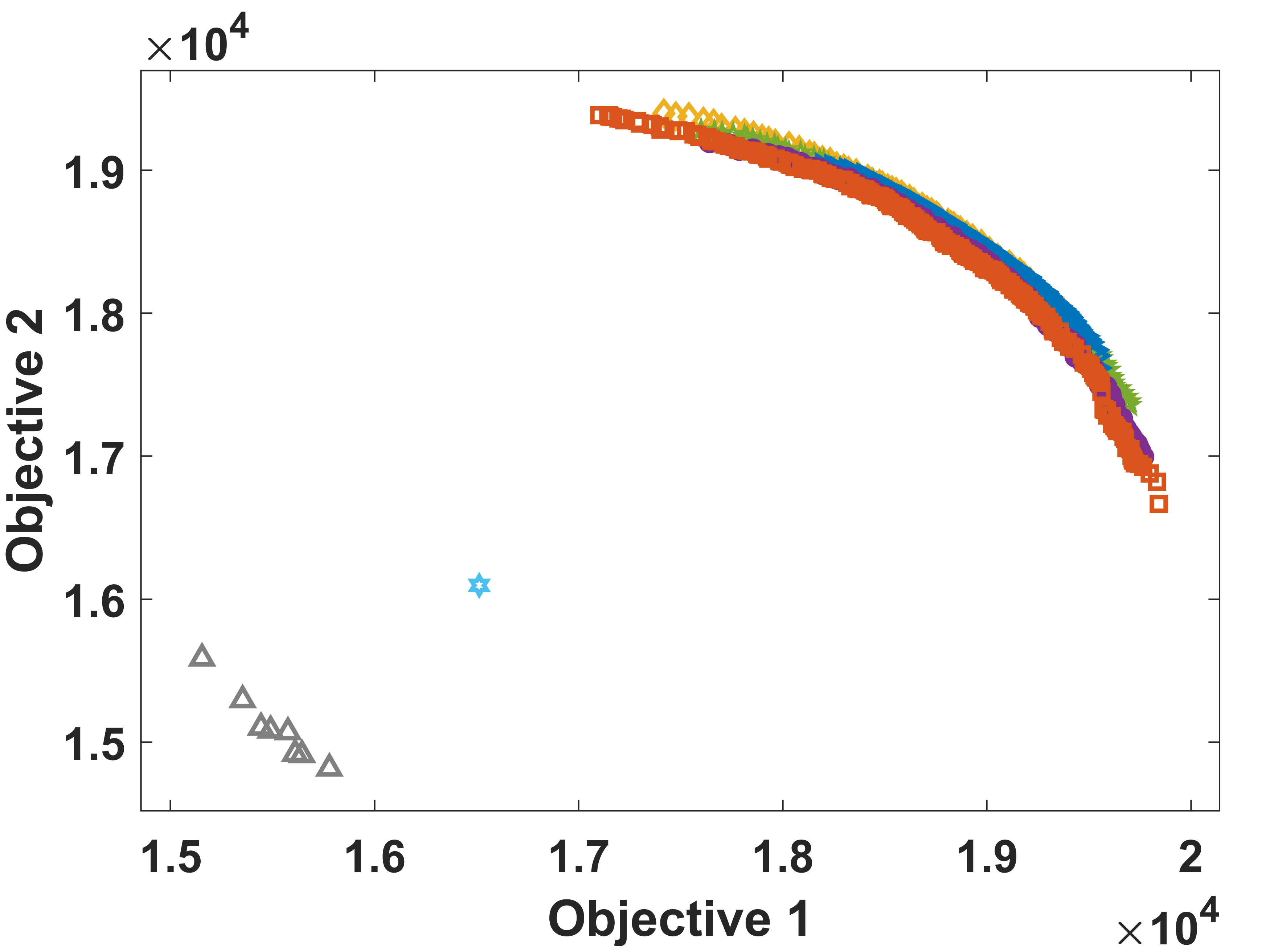} }
	\subfigure[Knapsack on eval.=\(5\times10^6\)]{
		\includegraphics[width=1.6in]{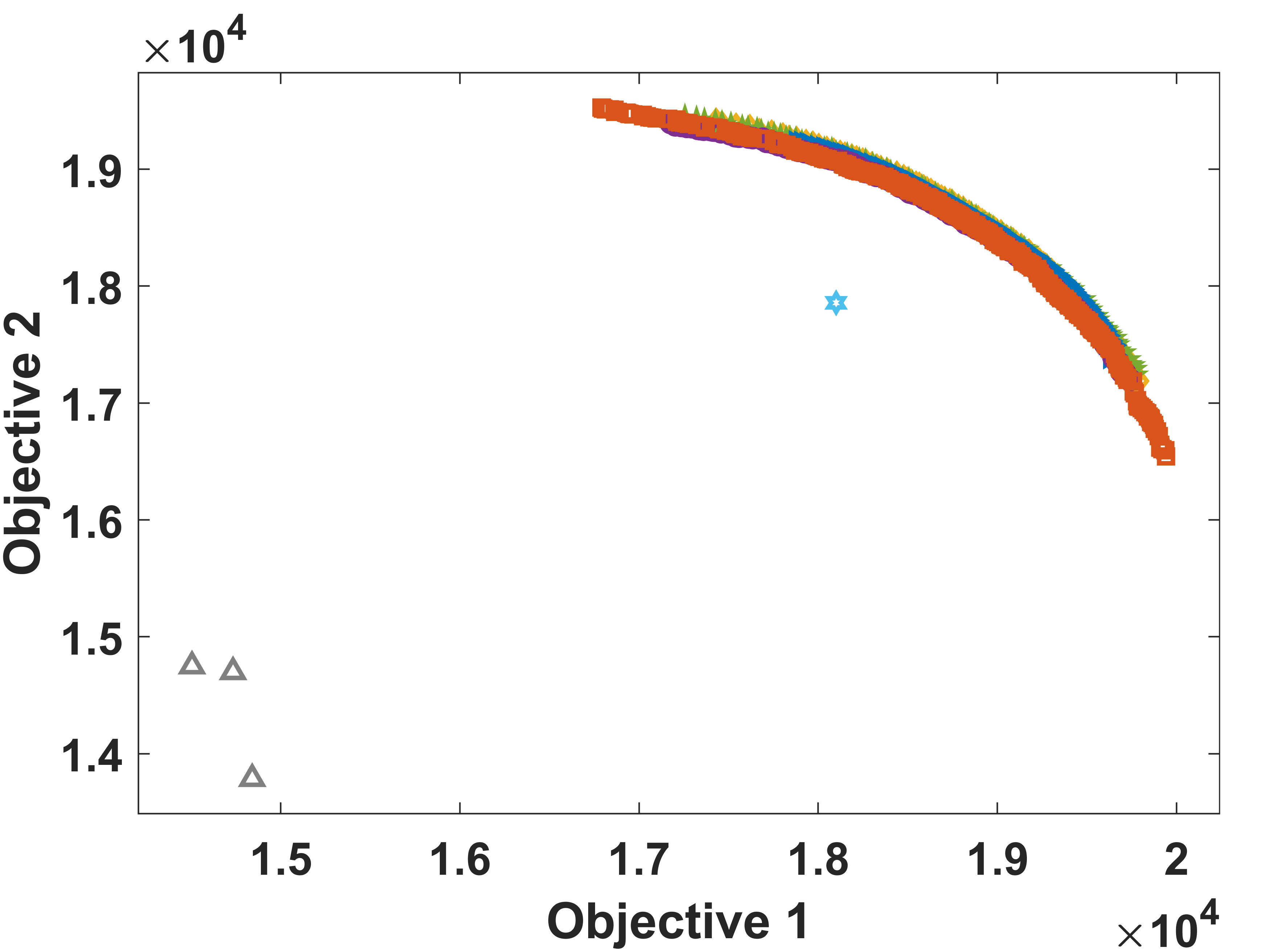} }
	\caption{The final archive sets are the results of the seven algorithms on 4 different settings of the third experiment from a single run. This particular run is associated with the result closest to the average HV value.}
	\label{Figure11}
\end{figure*}

\subsection{Comparison at different numbers of evaluations}
Table~\ref{Table3} presents the mean and standard deviation of the HV values obtained by the seven algorithms on the 16 test instances. Fig.~\ref{Figure11} shows the final solution set from a single run whose result is closest to the average HV value for this problem.

We consider the influence of different numbers of evaluations on algorithm performance under the same general settings as in the first experiment. As shown in Table~\ref{Table3} and Fig.~\ref{Figure11}, a clear trend can be observed, when the number of evaluations is small (Fig.~\ref{Figure11}(a)), the performance of VS-RLS decreases significantly, whereas when the number of evaluations reaches \(5\times10^5\) (Fig.~\ref{Figure11}(b)) or more (Figs.~\ref{Figure11}(c) and (d)), VS-RLS achieves the best performance across all problems. This result explains why VS-RLS performs poorly with fewer evaluations. When the total number of evaluations is limited to \(1\times10^5\), the algorithm may exhaust its computational resources during the early stages of the search. For example, in the Knapsack problem with \( D \)=500 and $T_{vl}$=1000, the worst-case scenario occurs when the algorithm gradually expands its stepsize from 1-bit to 500-bit flips without finding an improved solution. In such a case, 1000 iterations would require approximately \(5\times10^5\) evaluations, meaning that the algorithm may run out of evaluations before entering the exploitation phase. In addition, for the Knapsack problem, when a larger number of evaluations is provided than in the first experiment, the algorithm continues to show a gradual convergence trend. Specifically, when the number of evaluations reaches \(5\times10^6\) (Fig.~\ref{Figure11}(d)), VS-RLS achieves convergence comparable to that of MOEAs while maintaining higher solution diversity.

\subsection{Parameter Sensitivity Analysis and Ablation Study}
We conducted experiments on the Knapsack problem with a dimension of 500 and a total of 500,000 evaluations. This configuration was selected because, in the worst case, the algorithm may evaluate all \( D \) decision variables in each iteration. Running \( T_{vl}=1000 \) iterations therefore requires 500,000 evaluations, providing a fair basis for parameter analysis and ablation studies.

This experiment aims to evaluate two key components of the algorithm: \( V_C \), which controls the stepsize threshold during the exploitation phase, and \( T_{vl} \), which determines the transition point between the early and later phases of evolution. The experiments are divided into the following groups:
\subsubsection{\textbf{Baseline Group}} 
The baseline uses \( T_{vl}=1000 \) and \( V_C=3 \), as established in previous experiments. 
\subsubsection{\textbf{Stepsize Variation Group}} 
To investigate the effect of different variable stepsizes, we tested two additional settings: \( V_C=1 \) and \( V_C=5 \), corresponding to smaller and larger stepsize threshold adjustments, respectively.
\subsubsection{\textbf{Exploration Phase Ablation Group}} 
To assess the role of large stepsize threshold in the exploration phase, we created an ablation variant that removes this operation. This is equivalent to setting \( T_{vl}=0 \) while keeping \( V_C=3 \).

\begin{figure}[t] 
\centering  
\includegraphics[width=2.9in]{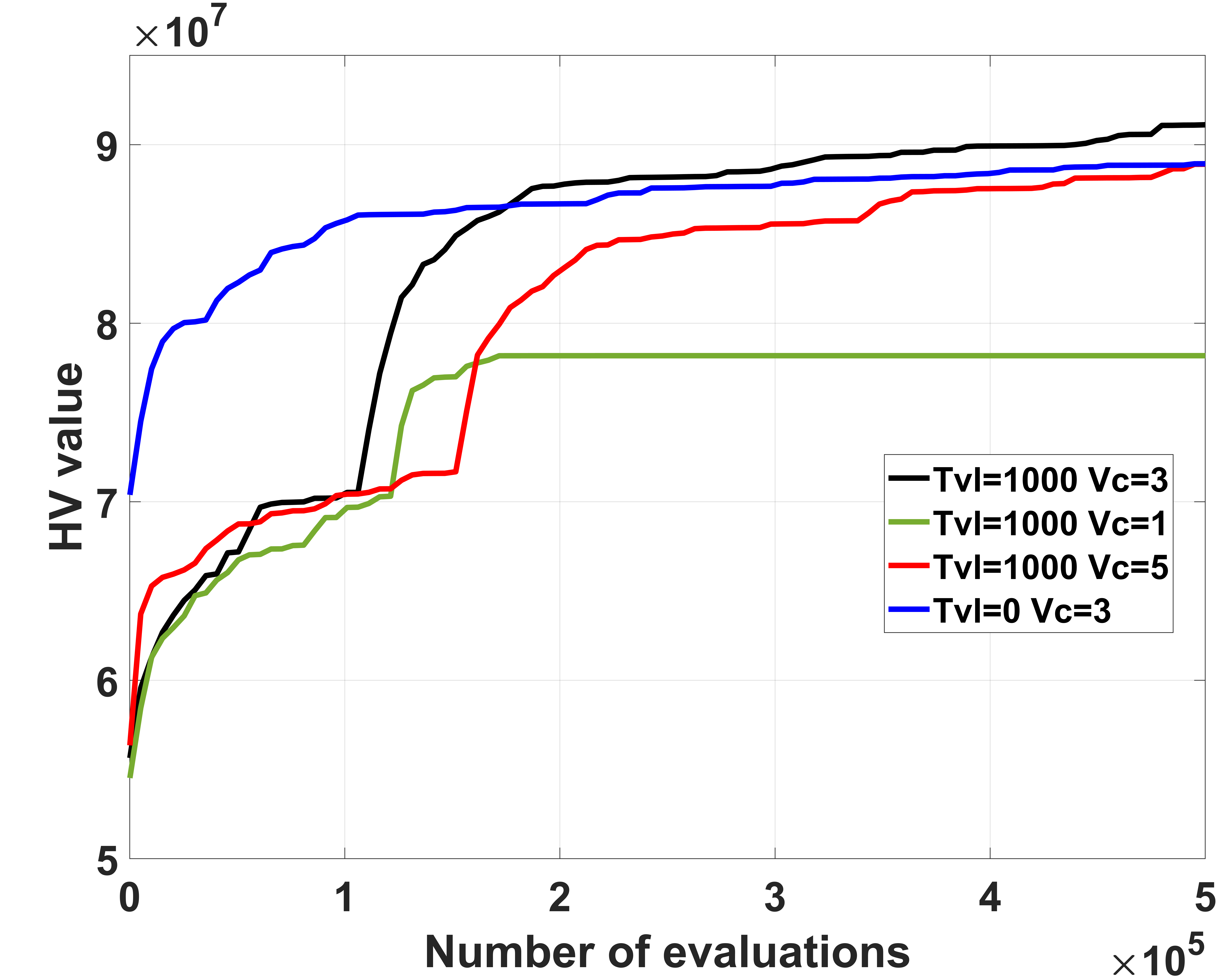} 
\caption{HV convergence curves of VS-RLS with different parameter settings. 
The vertical axis denotes the HV value, and the horizontal axis denotes the number of evaluations.}
\label{Figure8} 
\end{figure} 

Fig.~\ref{Figure8} presents the HV convergence curves under the original setting and three comparative configurations. When comparing the cases of \( T_{vl}=1000 \) with \( V_C=1 \) and \( V_C=5 \), the convergence trends are similar to those of the original configuration. After approximately 1,000 iterations, reducing the variable stepsize accelerates convergence, leading to a noticeable increase in HV. However, the overall performance under \( V_C=1 \) and \( V_C=5 \) is inferior to that of the baseline. This difference can be attributed to the varying search regions of decision variables. When \( V_C=1 \), the search region becomes too narrow, resulting in limited diversity and premature convergence to local optima. Conversely, when \( V_C=5 \), the search region becomes excessively large, causing many evaluations to be wasted. In contrast, the choice of \( V_C=3 \) represents a balanced trade-off between convergence speed and solution diversity.

In the ablation experiment, when \( T_{vl}=0 \) and \( V_C=3 \), the algorithm exhibits faster progress. However, this configuration depends heavily on the quality of the initial solution, making it difficult to accumulate high-quality solutions during this phase and weakening the search performance in the subsequent exploitation phase. This result demonstrates that a large stepsize is necessary during the exploration phase. 

\section{CONCLUSION}
In this paper, we studied the use of variable search stepsize in randomized local search for multi-objective combinatorial optimization. We showed that a simple mechanism that adjusts the size of neighbourhood exploration can lead to clear performance improvements across a wide range of problems. The proposed VS-RLS method begins with a broad search that encourages exploration of distant regions in the search space and then shifts to smaller and more focused regions as the search progresses. The empirical results on four representative MOCOPs show that VS-RLS is competitive against both randomized local search with fixed neighbourhoods and several well-established MOEAs. 

There are several directions for future work. One possibility is to integrate the variable stepsize mechanism into population-based algorithms or hybrid frameworks, allowing global and local search processes to complement each other. Another direction is to investigate more advanced adaptation strategies, for example data-driven or learning-based approaches that adjust the stepsize based on feedback collected during the search.


\bibliographystyle{IEEEtran}
\bibliography{myref}

@article{chen2023weights,
  title={The weights can be harmful: Pareto search versus weighted search in multi-objective search-based software engineering},
  author={Chen, Tao and Li, Miqing},
  journal={ACM Transactions on Software Engineering and Methodology},
  volume={32},
  number={1},
  pages={1--40},
  year={2023},
  publisher={ACM New York, NY}
}

@article{li2019quality,
  title={Quality evaluation of solution sets in multiobjective optimisation: A survey},
  author={Li, Miqing and Yao, Xin},
  journal={ACM Computing Surveys (CSUR)},
  volume={52},
  number={2},
  pages={1--38},
  year={2019},
  publisher={ACM New York, NY, USA}
}

@inproceedings{liang2023non,
  title={Non-elitist evolutionary multi-objective optimisation: Proof-of-principle results},
  author={Liang, Zimin and Li, Miqing and Lehre, Per Kristian},
  booktitle={Proceedings of the Companion Conference on Genetic and Evolutionary Computation},
  pages={383--386},
  year={2023}
}

@article{chu2024improving,
  title={Improving decomposition-based MOEAs for combinatorial optimisation by intensifying corner weights},
  author={Chu, Xiaochen and Han, Xiaofeng and Zhang, Maorui and Li, Miqing},
  journal={Swarm and Evolutionary Computation},
  volume={91},
  pages={101722},
  year={2024},
  publisher={Elsevier}
}

@article{zhu2015evolutionary,
  title={Evolutionary multi-objective workflow scheduling in cloud},
  author={Zhu, Zhaomeng and Zhang, Gongxuan and Li, Miqing and Liu, Xiaohui},
  journal={IEEE Transactions on parallel and distributed Systems},
  volume={27},
  number={5},
  pages={1344--1357},
  year={2015},
  publisher={IEEE}
}

@article{hierons2020many,
  title={Many-objective test suite generation for software product lines},
  author={Hierons, Robert M and Li, Miqing and Liu, Xiaohui and Parejo, Jose Antonio and Segura, Sergio and Yao, Xin},
  journal={ACM Transactions on Software Engineering and Methodology},
  volume={29},
  number={1},
  pages={1--46},
  year={2020},
  publisher={ACM New York, NY, USA}
}

@inproceedings{liang2026random,
title     = {Random is faster than systematic in multi-objective local search},
  author    = {Liang, Zimin and Li, Miqing},
  booktitle = {Proceedings of the 40th {AAAI} Conference on Artificial Intelligence},
  year      = {2026}
}

@inproceedings{bian2024archive,
  title        = {An archive can bring provable speed-ups in multi-objective evolutionary algorithms},
  author       = {Bian, Chao and Ren, Shuo and Li, Miqing and Qian, Chao},
  booktitle    = {Proceedings of the 33rd International Joint Conference on Artificial Intelligence ({IJCAI-24})},
  pages        = {6905--6913},
  year         = {2024}
}

@inproceedings{ren2026not,
	title = {Not Just for Archiving: Provable Benefits of Reusing the Archive in Evolutionary Multi-objective Optimization},
	author = {Ren, S. and Liang, Z. and Li, M. and Qian, C.},
	booktitle = {Proceedings of the AAAI Conference on Artificial Intelligence},
	year = {2026}
}

@inproceedings{li2025combinatorial,
  title={{Combinatorial Optimisation Can be Different from Continuous Optimisation for MOEAs}},
  author={Li, Miqing},
  booktitle={Proceedings of the Genetic and Evolutionary Computation Conference Companion},
  pages={1167--1181},
  year={2025}
}

@article{KP,
	title={Multiobjective evolutionary algorithms: a comparative case study and the strength Pareto approach},
	author={Zitzler, Eckart and Thiele, Lothar},
	journal={IEEE transactions on Evolutionary Computation},
	volume={3},
	number={4},
	pages={257--271},
	year={1999},
	publisher={IEEE}
}

@article{50,
	title={How to evaluate solutions in Pareto-based search-based software engineering: A critical review and methodological guidance},
	author={Li, Miqing and Chen, Tao and Yao, Xin},
	journal={IEEE Transactions on Software Engineering},
	volume={48},
	number={5},
	pages={1771--1799},
	year={2020},
	publisher={IEEE}
}

@inproceedings{NK,
	title={{Effects of elitism and population climbing on multiobjective MNK-landscapes}},
	author={Aguirre, Hern{\'a}n E and Tanaka, Kiyoshi},
	booktitle={Proceedings of the 2004 Congress on Evolutionary Computation (IEEE Cat. No. 04TH8753)},
	volume={1},
	pages={449--456},
	year={2004},
	organization={IEEE}
}

@inproceedings{QAP,
	title={Instance generators and test suites for the multiobjective quadratic assignment problem},
	author={Knowles, Joshua and Corne, David},
	booktitle={International Conference on Evolutionary Multi-Criterion Optimization},
	pages={295--310},
	year={2003},
	organization={Springer}
}

@article{TSP,
	title={A study of global convexity for a multiple objective travelling salesman problem},
	author={Ribeiro, Celso C and Hansen, Pierre and Borges, Pedro Castro and Hansen, Michael Pilegaard},
	journal={Essays and surveys in metaheuristics},
	pages={129--150},
	year={2002},
	publisher={Springer}
}

@incollection{PLS,
	title={Pareto local optimum sets in the biobjective traveling salesman problem: An experimental study},
	author={Paquete, Luis and Chiarandini, Marco and St{\"u}tzle, Thomas},
	booktitle={Metaheuristics for multiobjective optimisation},
	pages={177--199},
	year={2004},
	publisher={Springer}
}

@incollection{MOEADTSP,
	title={{Comparison between MOEA/D and NSGA-II on the multi-objective travelling salesman problem}},
	author={Peng, Wei and Zhang, Qingfu and Li, Hui},
	booktitle={Multi-objective memetic algorithms},
	pages={309--324},
	year={2009},
	publisher={Springer}
}

@article{TSP1,
	title={Mapping the global structure of TSP fitness landscapes},
	author={Ochoa, Gabriela and Veerapen, Nadarajen},
	journal={Journal of Heuristics},
	volume={24},
	number={3},
	pages={265--294},
	year={2018},
	publisher={Springer}
}

@inproceedings{III1,
  title={Visualising the landscape of multi-objective problems using local optima networks},
  author={Fieldsend, Jonathan E and Alyahya, Khulood},
  booktitle={Proceedings of the Genetic and Evolutionary Computation Conference Companion},
  pages={1421--1429},
  year={2019}
}

@book{count1,
  title={Multicriteria optimization},
  author={Ehrgott, Matthias},
  volume={491},
  year={2005},
  publisher={Springer Science \& Business Media}
}

@article{count2,
  title={A genetic algorithm approach for multi-objective optimization of supply chain networks},
  author={Altiparmak, Fulya and Gen, Mitsuo and Lin, Lin and Paksoy, Turan},
  journal={Computers \& industrial engineering},
  volume={51},
  number={1},
  pages={196--215},
  year={2006},
  publisher={Elsevier}
}

@article{count3,
  title={Satellite constellation design tradeoffs using multiple-objective evolutionary computation},
  author={Ferringer, Matthew P and Spencer, David B},
  journal={Journal of spacecraft and rockets},
  volume={43},
  number={6},
  pages={1404--1411},
  year={2006}
}

@inproceedings{count6,
  title={{MOEAs are stuck in a different area at a time}},
  author={Li, Miqing and Han, Xiaofeng and Chu, Xiaochen},
  booktitle={Proceedings of the Genetic and Evolutionary Computation Conference},
  pages={303--311},
  year={2023}
}

@article{count7,
	title={{A fast and elitist multiobjective genetic algorithm: NSGA-II}},
	author={Deb, Kalyanmoy and Pratap, Amrit and Agarwal, Sameer and Meyarivan, TAMT},
	journal={IEEE transactions on evolutionary computation},
	volume={6},
	number={2},
	pages={182--197},
	year={2002},
	publisher={IEEE}
}

@article{count8,
	title={{MOEA/D: A multiobjective evolutionary algorithm based on decomposition}},
	author={Zhang, Qingfu and Li, Hui},
	journal={IEEE Transactions on evolutionary computation},
	volume={11},
	number={6},
	pages={712--731},
	year={2007},
	publisher={IEEE}
}

@article{count9,
	title={{SMS-EMOA: Multiobjective selection based on dominated hypervolume}},
	author={Beume, Nicola and Naujoks, Boris and Emmerich, Michael},
	journal={European Journal of Operational Research},
	volume={181},
	number={3},
	pages={1653--1669},
	year={2007},
	publisher={Elsevier}
}

@article{count10,
  title={A survey on evolutionary computation for complex continuous optimization},
  author={Zhan, Zhi-Hui and Shi, Lin and Tan, Kay Chen and Zhang, Jun},
  journal={Artificial Intelligence Review},
  volume={55},
  number={1},
  pages={59--110},
  year={2022},
  publisher={Springer}
}

@article{count12,
  title={A genetic algorithm based approach to solve multi-resource multi-objective knapsack problem for vegetable wholesalers in fuzzy environment},
  author={Changdar, Chiranjit and Pal, Rajat Kumar and Mahapatra, Ghanshaym Singha and Khan, Abhinandan},
  journal={Operational Research},
  volume={20},
  number={3},
  pages={1321--1352},
  year={2020},
  publisher={Springer}
}

@article{count13,
  title={{MOEA/D-ACO: A multiobjective evolutionary algorithm using decomposition and antcolony}},
  author={Ke, Liangjun and Zhang, Qingfu and Battiti, Roberto},
  journal={IEEE transactions on cybernetics},
  volume={43},
  number={6},
  pages={1845--1859},
  year={2013},
  publisher={IEEE}
}

@article{count14,
  title={{An effective method for solving multiple travelling salesman problem based on NSGA-II}},
  author={Shuai, Yang and Yunfeng, Shao and Kai, Zhang},
  journal={Systems Science \& Control Engineering},
  volume={7},
  number={2},
  pages={108--116},
  year={2019},
  publisher={Taylor \& Francis}
}

@article{count15,
  title={{Multi-objective optimization of reliability--redundancy allocation problem with cold-standby strategy using NSGA-II}},
  author={Ardakan, Mostafa Abouei and Rezvan, Mohammad Taghi},
  journal={Reliability Engineering \& System Safety},
  volume={172},
  pages={225--238},
  year={2018},
  publisher={Elsevier}
}

@article{count16,
  title={{Multi-objective optimization of buffer allocation for remanufacturing system based on TS-NSGAII hybrid algorithm}},
  author={Su, Chun and Shi, Yangmei and Dou, Jianping},
  journal={Journal of Cleaner Production},
  volume={166},
  pages={756--770},
  year={2017},
  publisher={Elsevier}
}

@article{count17,
  title={A hybrid intelligent model for order allocation planning in make-to-order manufacturing},
  author={Guo, ZX and Wong, Wai Keung and Leung, Sunney Yung-Sun},
  journal={Applied Soft Computing},
  volume={13},
  number={3},
  pages={1376--1390},
  year={2013},
  publisher={Elsevier}
}

@inproceedings{count20,
  title={{NSGA-Net: neural architecture search using multi-objective genetic algorithm}},
  author={Lu, Zhichao and Whalen, Ian and Boddeti, Vishnu and Dhebar, Yashesh and Deb, Kalyanmoy and Goodman, Erik and Banzhaf, Wolfgang},
  booktitle={Proceedings of the genetic and evolutionary computation conference},
  pages={419--427},
  year={2019}
}

@article{count23,
  title={On the performance of multiple-objective genetic local search on the 0/1 knapsack problem-a comparative experiment},
  author={Jaszkiewicz, Andrzej},
  journal={IEEE Transactions on Evolutionary Computation},
  volume={6},
  number={4},
  pages={402--412},
  year={2002},
  publisher={IEEE}
}

@article{count25,
  title={Improving the anytime behavior of two-phase local search},
  author={Dubois-Lacoste, J{\'e}r{\'e}mie and L{\'o}pez-Ib{\'a}{\~n}ez, Manuel and St{\"u}tzle, Thomas},
  journal={Annals of mathematics and artificial intelligence},
  volume={61},
  pages={125--154},
  year={2011},
  publisher={Springer}
}

@article{count26,
  title={On dominance-based multiobjective local search: design, implementation and experimental analysis on scheduling and traveling salesman problems},
  author={Liefooghe, Arnaud and Humeau, J{\'e}r{\'e}mie and Mesmoudi, Salma and Jourdan, Laetitia and Talbi, El-Ghazali},
  journal={Journal of Heuristics},
  volume={18},
  pages={317--352},
  year={2012},
  publisher={Springer}
}

@inproceedings{count27,
  title={Clusters of non-dominated solutions in multiobjective combinatorial optimization: An experimental analysis},
  author={Paquete, Lu{\'\i}s and St{\"u}tzle, Thomas},
  booktitle={Multiobjective Programming and Goal Programming: Theoretical Results and Practical Applications},
  pages={69--77},
  year={2009},
  organization={Springer}
}

@article{count28,
  title={Effective anytime algorithm for multiobjective combinatorial optimization problems},
  author={Dom{\'\i}nguez-R{\'\i}os, Miguel {\'A}ngel and Chicano, Francisco and Alba, Enrique},
  journal={Information Sciences},
  volume={565},
  pages={210--228},
  year={2021},
  publisher={Elsevier}
}

@article{count29,
  title={{jMetal: A Java framework for multi-objective optimization}},
  author={Durillo, Juan J and Nebro, Antonio J},
  journal={Advances in engineering software},
  volume={42},
  number={10},
  pages={760--771},
  year={2011},
  publisher={Elsevier}
}

@inproceedings{count34,
  title={{Empirical comparison between MOEAs and local search on multi-objective combinatorial optimisation problems}},
  author={Li, Miqing and Han, Xiaofeng and Chu, Xiaochen and Liang, Zimin},
  booktitle={Proceedings of the Genetic and Evolutionary Computation Conference},
  pages={547--556},
  year={2024}
}

@article{count35,
	title={Running time analysis of multiobjective evolutionary algorithms on pseudo-boolean functions},
	author={Laumanns, Marco and Thiele, Lothar and Zitzler, Eckart},
	journal={IEEE Transactions on Evolutionary Computation},
	volume={8},
	number={2},
	pages={170--182},
	year={2004},
	publisher={IEEE}
}

@inproceedings{1Relatedwork1,
  title={{Global vs local search on multi-objective NK-landscapes: contrasting the impact of problem features}},
  author={Daolio, Fabio and Liefooghe, Arnaud and Verel, S{\'e}bastien and Aguirre, Hern{\'a}n and Tanaka, Kiyoshi},
  booktitle={proceedings of the 2015 Annual Conference on Genetic and Evolutionary Computation},
  pages={369--376},
  year={2015}
}

@inproceedings{1Relatedwork4,
  title={A hybrid evolutionary approach for multicriteria optimization problems: Application to the flow shop},
  author={Talbi, El-Ghazali and Rahoual, Malek and Mabed, Mohamed Hakim and Dhaenens, Clarisse},
  booktitle={Evolutionary Multi-Criterion Optimization: First International Conference, EMO 2001 Zurich, Switzerland, March 7--9, 2001 Proceedings 1},
  pages={416--428},
  year={2001},
  organization={Springer}
}

@article{2Relatedwork14,
  title={Random bit climbers on multiobjective MNK-landscapes: effects of memory and population climbing},
  author={Aguirre, Hernan and Tanaka, Kiyoshi},
  journal={IEICE transactions on fundamentals of electronics, communications and computer sciences},
  volume={88},
  number={1},
  pages={334--345},
  year={2005},
  publisher={The Institute of Electronics, Information and Communication Engineers}
}

@inproceedings{1GSEMO,
  title={The (1+($\lambda$, $\lambda$)) global SEMO algorithm},
  author={Doerr, Benjamin and Hadri, Omar El and Pinard, Adrien},
  booktitle={Proceedings of the Genetic and Evolutionary Computation Conference},
  pages={520--528},
  year={2022}
}

@inproceedings{BCGSEMO,
  title={A block-coordinate descent emo algorithm: Theoretical and empirical analysis},
  author={Doerr, Benjamin and Knowles, Joshua and Neumann, Aneta and Neumann, Frank},
  booktitle={Proceedings of the Genetic and Evolutionary Computation Conference},
  pages={493--501},
  year={2024}
}

@article{PLS2,
  title={Two-phase Pareto local search for the biobjective traveling salesman problem},
  author={Lust, Thibaut and Teghem, Jacques},
  journal={Journal of Heuristics},
  volume={16},
  number={3},
  pages={475--510},
  year={2010},
  publisher={Springer}
}

@article{PLS8,
  title={Anytime Pareto local search},
  author={Dubois-Lacoste, J{\'e}r{\'e}mie and L{\'o}pez-Ib{\'a}{\~n}ez, Manuel and St{\"u}tzle, Thomas},
  journal={European journal of operational research},
  volume={243},
  number={2},
  pages={369--385},
  year={2015},
  publisher={Elsevier}
}

@inproceedings{PLS9,
  title={Improving Neighborhood Exploration Mechanism to Speed up PLS},
  author={Kang, Yuhao and Shi, Jialong and Sun, Jianyong and Fan, Ye},
  booktitle={Proceedings of the genetic and evolutionary computation conference},
  pages={688--694},
  year={2023}
}

@article{PLS17,
  title={On local search for bi-objective knapsack problems},
  author={Liefooghe, Arnaud and Paquete, Lu{\'\i}s and Figueira, Jos{\'e} Rui},
  journal={Evolutionary computation},
  volume={21},
  number={1},
  pages={179--196},
  year={2013},
  publisher={MIT Press One Rogers Street, Cambridge, MA 02142-1209, USA journals-info~…}
}

@inproceedings{PLS22,
  title={Learning to balance exploration and exploitation in pareto local search for multi-objective combinatorial optimization},
  author={Zhang, Haotian and Shi, Jialong and Sun, Jianyong and Xu, Zongben},
  booktitle={Proceedings of the genetic and evolutionary computation conference companion},
  pages={383--386},
  year={2022}
}

@article{PLS23,
  title={New techniques to improve neighborhood exploration in pareto local search},
  author={Kang, Yuhao and Shi, Jialong and Sun, Jianyong and Zhang, Qingfu and Fan, Ye},
  journal={Expert Systems with Applications},
  volume={254},
  pages={124296},
  year={2024},
  publisher={Elsevier}
}

@inproceedings{NSEMO6,
  title={An analysis on recombination in multi-objective evolutionary optimization},
  author={Qian, Chao and Yu, Yang and Zhou, Zhi-Hua},
  booktitle={Proceedings of the 13th annual conference on Genetic and evolutionary computation},
  pages={2051--2058},
  year={2011}
}

@incollection{NSEMO10,
  title={Iterated local search: Framework and applications},
  author={Louren{\c{c}}o, Helena Ramalhinho and Martin, Olivier C and St{\"u}tzle, Thomas},
  booktitle={Handbook of metaheuristics},
  pages={129--168},
  year={2018},
  publisher={Springer}
}

@inproceedings{NSEMO11,
  title={Multi-objective Order Reduction Problem Solving with Restart Meta-heuristic Implementation},
  author={Ryzhikov, Ivan and Brester, Christina and Semenkin, Eugene},
  booktitle={ICINCO (1)},
  pages={270--278},
  year={2017}
}

@article{NAS1,
  title={An evolutionary multi-objective neural architecture search approach to advancing cognitive diagnosis in intelligent education},
  author={Yang, Shangshang and Ma, Haiping and Bi, Ying and Tian, Ye and Zhang, Limiao and Jin, Yaochu and Zhang, Xingyi},
  journal={IEEE Transactions on Evolutionary Computation},
  year={2024},
  publisher={IEEE}
}

@article{NAS2,
  title={{MedNAS}: Multiscale training-free neural architecture search for medical image analysis},
  author={Wang, Yan and Zhen, Liangli and Zhang, Jianwei and Li, Miqing and Zhang, Lei and Wang, Zizhou and Feng, Yangqin and Xue, Yu and Wang, Xiao and Chen, Zheng and others},
  journal={IEEE Transactions on Evolutionary Computation},
  volume={28},
  number={3},
  pages={668--681},
  year={2024},
  publisher={IEEE}
}

@article{NAS3,
  title={Surrogate-assisted evolutionary multiobjective neural architecture search based on transfer stacking and knowledge distillation},
  author={Lyu, Kuangda and Li, Hao and Gong, Maoguo and Xing, Lining and Qin, Alex Kai},
  journal={IEEE Transactions on Evolutionary Computation},
  volume={28},
  number={3},
  pages={608--622},
  year={2023},
  publisher={IEEE}
}

@inproceedings{TSPcanshu,
  title={Techniques for highly multiobjective optimisation: some nondominated points are better than others},
  author={Corne, David W and Knowles, Joshua D},
  booktitle={Proceedings of the 9th annual conference on Genetic and evolutionary computation},
  pages={773--780},
  year={2007}
}

@article{QAPcanshu,
  title={Comparison of iterative searches for the quadratic assignment problem},
  author={Taillard, Eric D},
  journal={Location science},
  volume={3},
  number={2},
  pages={87--105},
  year={1995},
  publisher={Elsevier}
}

@book{canshu,
  title={Introduction to evolutionary computing},
  author={Eiben, Agoston E and Smith, James E},
  year={2015},
  publisher={Springer}
}

@inproceedings{zimin2,
  title={Pareto local search is competitive with evolutionary algorithms for multi-Objective neural architecture search},
  author={Phan, Quan Minh and Luong, Ngoc Hoang},
  booktitle={Proceedings of the Genetic and Evolutionary Computation Conference},
  pages={348--356},
  year={2023}
}

@article{RW2,
  title={Stochastic local search algorithms for multiobjective combinatorial optimization: A review},
  author={Paquete, Luis and St{\"u}tzle, Thomas},
  journal={Handbook of approximation algorithms and metaheuristics},
  pages={411--425},
  year={2018},
  publisher={Chapman and Hall/CRC}
}

@article{RW44,
  title={Towards Landscape Analyses to Inform the Design of Hybrid Local Search for the Multiobjective Quadratic Assignment Problem.},
  author={Knowles, Joshua D and Corne, David and others},
  journal={HIS},
  volume={87},
  pages={271--279},
  year={2002},
  publisher={Citeseer}
}

@article{DMOLS2,
  title={On local optima in multiobjective combinatorial optimization problems},
  author={Paquete, Luis and Schiavinotto, Tommaso and St{\"u}tzle, Thomas},
  journal={Annals of Operations Research},
  volume={156},
  pages={83--97},
  year={2007},
  publisher={Springer}
}

@inproceedings{DMOLS4,
  title={Design of multi-objective evolutionary algorithms: Application to the flow-shop scheduling problem},
  author={Basseur, Matthieu and Seynhaeve, Franck and Talbi, El-ghazali},
  booktitle={Proceedings of the 2002 Congress on Evolutionary Computation. CEC'02 (Cat. No. 02TH8600)},
  volume={2},
  pages={1151--1156},
  year={2002},
  organization={IEEE}
}

@book{SCAandPar,
  title={Design of heuristic algorithms for hard optimization: with python codes for the travelling salesman problem},
  author={Taillard, {\'E}ric D},
  year={2023},
  publisher={Springer Nature}
}

@inproceedings{Bit1,
  title={Comparing global and local mutations on bit strings},
  author={Doerr, Benjamin and Jansen, Thomas and Klein, Christian},
  booktitle={Proceedings of the 10th annual conference on Genetic and evolutionary computation},
  pages={929--936},
  year={2008}
}

@inproceedings{Liang2025,
author = {Liang, Zimin and Li, Miqing},
title = {On the Problem Characteristics of Multi-objective Pseudo-Boolean Functions in Runtime Analysis},
year = {2025},
booktitle = {Foundations of Genetic Algorithms},
pages = {166–177},
numpages = {12},
}

@inproceedings{deng_runtime_2024,
	title = {Runtime {analysis} for {state}-of-the-{art} {multi}-objective {evolutionary} {algorithms} on the {subset} {selection} {problem}},
	isbn = {978-3-031-70070-5},
	booktitle = {Parallel {Problem} {Solving} from {Nature}},
	author = {Deng, R. and Zheng, W. and Li, M. and Liu, J. and Doerr, B.},
	year = {2024},
	pages = {264--279}
}

@inproceedings{Qian2025,
author = {Qian, Chao},
title = {Pareto Optimization for Subset Selection: Theories and Practical Algorithms},
year = {2025},
isbn = {9798400714641},
publisher = {Association for Computing Machinery},
booktitle = {Proceedings of the Genetic and Evolutionary Computation Conference},
pages = {1592–1616},
numpages = {25},
}

@inproceedings{bian_robust_2022,
	title = {Robust {Subset} {Selection} by {Greedy} and {Evolutionary} {Pareto} {Optimization}},
	isbn = {978-1-956792-00-3},
	booktitle = {{International} {Joint} {Conference} on {Artificial} {Intelligence}},
	author = {Bian, Chao and Zhou, Yawen and Qian, Chao},
	year = {2022},
	pages = {4726--4732},
}

@article{do_rigorous_2023,
	title = {Rigorous Runtime Analysis of {MOEA}/{D} for Solving Multi-Objective Minimum Weight Base Problems},
	volume = {36},
	journal = {Advances in Neural Information Processing Systems},
	author = {Do, Anh Viet and Neumann, Aneta and Neumann, Frank and Sutton, Andrew},
	year = {2023},
	pages = {36434--36448},
}

@article{neumann_expected_2007,
	title = {Expected runtimes of a simple evolutionary algorithm for the multi-objective minimum spanning tree problem},
	volume = {181},
	issn = {0377-2217},
	number = {3},
	journal = {European Journal of Operational Research},
	author = {Neumann, Frank},
	year = {2007},
	pages = {1620--1629},
}

@inproceedings{zheng_first_2022,
	title = {A {first} {mathematical} {runtime} {analysis} of the {non}-dominated {sorting} {genetic} {algorithm} {II} ({NSGA}-{II})},
	volume = {36},
	issn = {2374-3468},
	number = {9},
	booktitle = {AAAI Conference on Artificial Intelligence},
	author = {Zheng, W. and Liu, Y. and Doerr, B.},
	year = {2022},
	pages = {10408--10416}
}

@article{doerr_first_2023,
	title = {A {first} {runtime} {analysis} of the {NSGA}-{II} on a {multimodal} {problem}},
	volume = {27},
	issn = {1941-0026},
	number = {5},
	journal = {Evolutionary Computation},
	author = {Doerr, B. and Qu, Z.},
	year = {2023},
	pages = {1288--1297}
}

@article{bian_stochastic_2025,
	title = {Stochastic population update can provably be helpful in multi-objective evolutionary algorithms},
	volume = {341},
	issn = {0004-3702},
	journal = {Artificial Intelligence},
	author = {Bian, Chao and Zhou, Yawen and Li, Miqing and Qian, Chao},
	year = {2025},
	pages = {104308},
}

@inproceedings{zheng_how_2024,
	title = {How to {use} the {metropolis} {algorithm} for {multi}-{objective} {optimization}?},
	volume = {38},
	issn = {2374-3468},
	language = {en},
	number = {18},
	booktitle = {AAAI Conference on Artificial Intelligence},
	author = {Zheng, W. and Li, M. and Deng, R. and Doerr, B.},
	year = {2024},
	pages = {20883--20891}
}

@inproceedings{doerr_proven_2024,
	title = {Proven {runtime} {guarantees} for {how} the {MOEA}/{D}: {computes} the {Pareto} {front} from the {subproblem} {solutions}},
	isbn = {978-3-031-70071-2},
	booktitle = {Parallel {Problem} {Solving} from {Nature}},
	author = {Doerr, B. and Krejca, M.S. and Weeks, N.},
	year = {2024},
	pages = {197--212}
}

@article{ren2025stochastic,
	title = {Stochastic Population Update Provably Needs An Archive in Evolutionary Multi-objective Optimization},
	author = {Ren, S. and Liang, Z. and Li, M. and Qian, C.},
	journal = {arXiv preprint arXiv:2501.16735},
	year = {2025}
}

@inproceedings{TSP2,
  title={A hybrid genetic algorithm for the traveling salesman problem using generalized partition crossover},
  author={Whitley, Darrell and Hains, Doug and Howe, Adele},
  booktitle={International Conference on Parallel Problem Solving from Nature},
  pages={566--575},
  year={2010},
  organization={Springer}
}
\end{document}